\documentclass{article}

    \PassOptionsToPackage{numbers, compress}{natbib}

\usepackage{amsmath,amsfonts,bm}

\def\eqref#1{equation~\ref{#1}}

\def\1{\bm{1}}

\def\eps{{\epsilon}}

\DeclareMathAlphabet{\mathsfit}{\encodingdefault}{\sfdefault}{m}{sl}
\SetMathAlphabet{\mathsfit}{bold}{\encodingdefault}{\sfdefault}{bx}{n}

\DeclareMathOperator*{\argmax}{arg\,max}

\newcommand{\Tr}[1]{\mathrm{Tr}\left\{#1\right\}}

\usepackage[T1]{fontenc}    %
\usepackage{hyperref}       %

\usepackage{url}            %
\usepackage{booktabs}       %
\usepackage{amsfonts}       %
\usepackage{nicefrac}       %
\usepackage{microtype}      %
\usepackage{xcolor}         %
\usepackage{colortbl}
\usepackage[normalem]{ulem}
\usepackage{algorithm}
\usepackage{algpseudocode}
\usepackage{wrapfig}
\usepackage{mathtools}
\usepackage{times}
\usepackage{epsfig}
\usepackage{graphicx}
\usepackage{amsmath}
\usepackage{amsthm}
\usepackage{amssymb}
\usepackage{bm}
\usepackage{svg}
\usepackage{caption}
\usepackage{subcaption}
\usepackage{multirow}
\usepackage{standalone}
\usepackage{url}
\usepackage{enumitem}
\usepackage{tcolorbox}
\usepackage{bbm}
\usepackage{soul}
\usepackage{listings}
\usepackage{tabularx}
\usepackage[capitalize]{cleveref}
\crefname{section}{\S}{\S\S}
\Crefname{section}{Section}{Sections}
\Crefname{table}{Tab.}{Tabs.}
\Crefname{equation}{Eq.}{Eqs.}
\Crefname{figure}{Fig.}{Figs.}
\Crefname{problem}{Prob.}{Probs.}
\crefname{table}{Tab.}{Tabs.}
\crefname{appendix}{\S}{\SS}
\crefname{definition}{Def.}{Defs.}
\Crefname{appendix}{Appendix}{Appendices}
\Crefname{lemma}{lemma}{lemma}
\usepackage{tikz}

\usetikzlibrary{arrows.meta}
\usetikzlibrary{plotmarks}
\usetikzlibrary{decorations}
\usepackage{pgfplots}
\pgfplotsset{compat=1.18} 
\usepackage{rotating}

\newcommand{\methodName}[0]{PolyJuice} %

\newcommand{\sepehr}[1]{}
\newcommand{\vishnu}[1]{}
\newcommand{\mashrur}[1]{}

\newcommand{\change}[1]{}

\newcommand{\Rt}[0]{red teaming}
\newcommand{\RT}[0]{Red teaming}
\newcommand{\Rtd}[0]{red-teaming}

\newcommand{\sd}[0]{SDv3.5}
\newcommand{\dev}{$\text{FLUX}_{\text{[dev]}}$}
\newcommand{\sch}{$\text{FLUX}_{\text{[sch]}}$}

\newcommand{\remark}[1]{
\begin{tcolorbox}[width=\textwidth,
                  colback=fillcolor7!30!green!20!white,
                  colframe=fillcolor7!70!black,
                  boxsep=3pt,
                  left=0pt,
                  right=0pt,
                  top=2pt,
                  bottom=2pt]%
\textbf{Takeaway.\quad} #1
\end{tcolorbox}
}

\theoremstyle{definition}
\theoremstyle{problem}

\newtheorem{problem}{Problem}

\newcount\gaussF

\definecolor{skyblue}{RGB}{232, 243, 255}

\newcommand{\hlight}[1]{%
    \tcbox[colback=green!30, colframe=green!30, arc=1mm, boxsep=0mm, boxrule=1mm, on line, leftrule=0mm, rightrule=0mm, right=1pt, left=1pt, top=0pt, bottom=0pt]{#1}
}
\newcommand{\highll}[1]{%
    \tcbox[colback=blue!10, colframe=blue!10, arc=1mm, boxsep=0mm, boxrule=1mm, on line, leftrule=0mm, rightrule=0mm, right=1pt, left=1pt, top=0pt, bottom=0pt]{#1}
}

\usetikzlibrary{positioning}
\usetikzlibrary{shapes.geometric}

\definecolor{sns1}{RGB}{20, 135, 198}
\definecolor{sns2}{RGB}{213, 94, 0}
\definecolor{sns3}{RGB}{2, 158, 115}
\definecolor{sns4}{RGB}{222, 143, 5}
\definecolor{sns5}{RGB}{204, 120, 188}
\colorlet{enc-col}{sns5!40}
\colorlet{cls-col}{black!10}

\definecolor{fillcolor}{HTML}{DEDEFF}  %
\definecolor{fillcolor2}{HTML}{CFD3D8} %
\definecolor{fillcolor4}{HTML}{f2f3f5} %
\definecolor{fillcolor5}{HTML}{f7f8fa} %
\definecolor{fillcolor3}{HTML}{FFE342} %
\definecolor{fillcolor6}{HTML}{ff8f8f} %
\definecolor{fillcolor7}{HTML}{54c45e} %
\definecolor{fillcolor8}{HTML}{ff9933} %
\definecolor{fillcolor9}{HTML}{2E6F40} %

\definecolor{fillcolor10}{HTML}{ffb343} %
\definecolor{fillcolor11}{HTML}{E86100} %
\definecolor{fillcolor12}{HTML}{6a5acd} %

\definecolor{colorbox}{HTML}{f21654} %

\definecolor{firered}{HTML}{A9423F} %
\definecolor{brickred}{HTML}{ad2020}

\definecolor{vibgreen}{HTML}{0d621e}

\definecolor{HotPink}{HTML}{FF69B4}
\hypersetup{
  colorlinks=true,      %
  urlcolor=HotPink,     %
  citecolor=green!70!black,    %
}

    \usepackage[final]{neurips/neurips_2025}

\usepackage[utf8]{inputenc} %
\usepackage[T1]{fontenc}    %
\usepackage{hyperref}       %
\usepackage{url}            %
\usepackage{booktabs}       %
\usepackage{amsfonts}       %
\usepackage{nicefrac}       %
\usepackage{microtype}      %
\usepackage{xcolor}         %

\title{\includegraphics[height=16pt, angle=15]{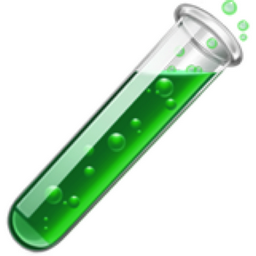}\textcolor{vibgreen}{\methodName{}} Makes It Real: Black-Box, Universal Red Teaming for Synthetic Image Detectors}

\author{
  Sepehr Dehdashtian$^{1*\dagger}$
  \quad Mashrur M. Morshed$^{1*}$
  \quad Jacob H. Seidman$^{2}$
  \quad Gaurav Bharaj$^{2}$ \\
  \textbf{Vishnu Naresh Boddeti}$^{1}$\\
  \\
  $^{1}$ Michigan State University \quad $^{2}$ Reality Defender
  \\
  {\small \texttt{\{sepehr,morshedm,vishnu\}@msu.edu}}
  \hfill
  {\small\texttt{\{jacob,gaurav\}@realitydefender.ai}}
}

\begin{document}

\maketitle

\vspace{-1em}
\vspace{-0.5em}
\begin{abstract}
\vspace{-0.5em}

Synthetic image detectors (SIDs) are a key defense against the risks posed by the growing realism of images from text-to-image (T2I) models. \RT{} improves SID's effectiveness by \emph{identifying} and \emph{exploiting} their failure modes via misclassified synthetic images.
However, existing \Rtd{} solutions (i) require white-box access to SIDs, which is infeasible for proprietary state-of-the-art detectors, and (ii) generate image-specific attacks through expensive online optimization.
To address these limitations, we propose \methodName{}, the first \emph{black-box}, image-agnostic \Rtd{} method for SIDs, based on an observed \emph{distribution shift} in the T2I latent space between samples correctly and incorrectly classified by the SID. 
\methodName{} generates attacks by (i) \emph{identifying} the direction of this shift through a lightweight offline process that only requires black-box access to the SID, and (ii) \emph{exploiting} this direction by universally steering all generated images towards the SID's failure modes.
\methodName{}-steered T2I models are significantly more effective at deceiving SIDs (up to 84\%) compared to their unsteered counterparts. We also show that the steering directions can be estimated efficiently at lower resolutions and transferred to higher resolutions using simple interpolation, reducing computational overhead. Finally, tuning SID models on \methodName{}-augmented datasets notably enhances the performance of the detectors (up to 30\%).

\end{abstract}

\begingroup
\renewcommand{\thefootnote}{*} %
\footnotetext{Equal contribution. \quad $\dagger$This work was done during an internship at Reality Defender.\\ 
\textcolor{white}{\ \ \ \ \ \ }
Project webpage: \url{https://sepehrdehdashtian.github.io/Papers/PolyJuice}} %
\endgroup

\begin{figure}[h]
    \centering
    \vspace{-0.5em}
    \begin{subfigure}[t]{0.65\linewidth}
        \includegraphics[width=\linewidth]{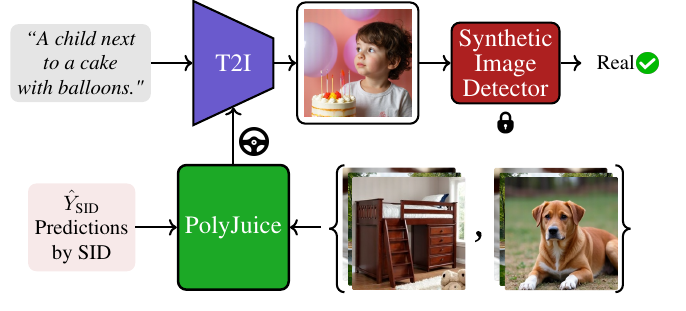}
        \caption{\label{fig:teaser:pj}}
    \end{subfigure}
    \begin{subfigure}[t]{0.29\linewidth}
        \includegraphics[width=\linewidth]{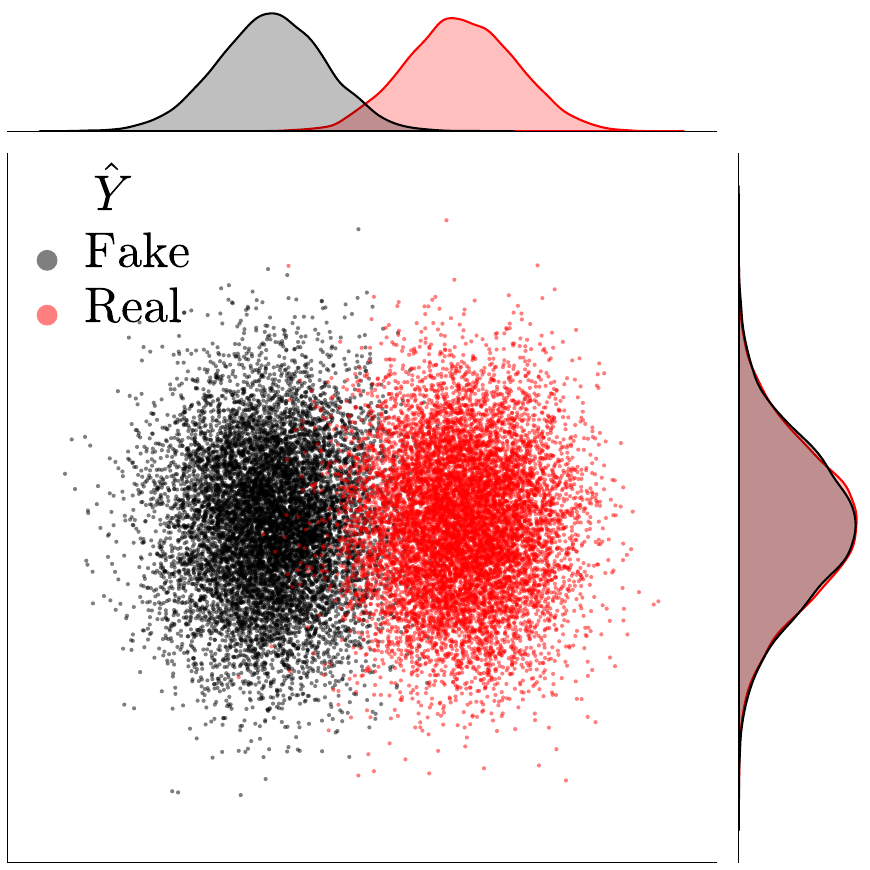}
        \caption{\hspace{3em}\label{fig:teaser:shift}}
    \end{subfigure}
    \caption{\footnotesize (a) \methodName{} \emph{steers} text-to-image (T2I) models to generate images that deceive a synthetic image detection (SID) model. (b) There exists a clearly observable shift between the distribution of the samples predicted as real versus those identified as fake, in the latent space of T2I models.}
    \label{fig:teaser}
    \vspace{-0.5em}
\end{figure}

\section{Introduction}

\textbf{Importance of synthetic image detection (SID) and \Rt{}.}\quad
Creating artificial content has never been easier. In recent years, rapid advances in text-to-image (T2I) generative models \citep{dhariwal2021diffusion,ramesh2021zero,rombach2022high,saharia2022photorealistic,midjourney, esser2024scaling,flux} have significantly blurred the line between real and fake visual media. To reinforce this line, it is necessary to improve the primary defense: synthetic image detection (SID) or deepfake detection (DFD) models \cite{corvi2023detection, ojha2023towards,koutlis2024leveraging,wang2020cnn,frank2020leveraging,shiohara2022detecting,sun2021improving}. One tried-and-true method for improving the robustness of critical machine learning systems is \emph{\Rt{}}: simulating attacks to discover failure modes. For example, \Rt{} is extensively used to expose vulnerabilities in biometric recognition systems \cite{hadid2015biometrics,zhu2021one,jia2022adv,li2023discrete}, and in large-language models (LLM) \cite{andriushchenko2024jailbreaking,li2024art,deng2023attack,perez-etal-2022-red}. Similarly, these attacks can help in improving SID models by discovering their blind spots.

\textbf{\RT{} as a way to improve SIDs.}\quad
An ideal \Rt{} for SID needs to have two features: \textbf{(i) Augmentation:} Fake detection datasets have not kept up with the rapidly improving field of image generation \citep{chandra2025deepfake}. To help improve these datasets, \Rt{} should discover a distribution of synthetic images capable of deceiving SID models. These images can be used to augment datasets with more challenging examples, helping to improve both evaluation and detection. \textbf{(ii) Black-Box Attacks}: State-of-the-art SID methods are mostly proprietary \cite{chandra2025deepfake} and provide limited access through APIs, making white-box attacks infeasible. Therefore, a practical \Rt{} method for SID must focus on black-box attacks that can be performed with model-provided responses only.

\textbf{Existing \Rtd{} methods and their limitations.}\quad
Recent years have seen the emergence of a family of attacks known as Unrestricted Adversarial (UA) attacks, that utilize powerful generative priors to directly create attack images capable of misleading a classifier \cite{song2018constructing}. However, existing UA attacks share some common limitations: \textbf{(L1) They perform white-box attacks}: Prior UA attacks \cite{xue2023diffusion,dai2024advdiff,chen2024diffusion} need access to the weights or gradients of the classifier to guide or modify the result of the generative model. These attacks can only be adapted to black-box detectors through the unreliable method of transferability from white-box settings. 
\textbf{(L2) They perform image-specific attacks}: Existing UA attacks need to optimize a \emph{per-image} perturbation or direction in the latent space of generative models, which adds a considerable computational overhead at inference time. Further, the optimization cost can grow exponentially with the image resolution.

\textbf{In this paper, we present \methodName{}}, a black-box method for generating UA attacks against SID models. For a given T2I generator and a target SID model, \methodName{} \emph{steers} the T2I generative process to discover images that successfully deceive the detector, as shown in \cref{fig:teaser:pj}.
\methodName{} identifies the steering direction from a set of generated images, by characterizing the subspace where the statistical dependence between their \emph{T2I latents} and their \emph{SID-predicted realness} is maximal.
Finding these directions is motivated by the key observation that there exists a distribution shift between the predicted real and fake samples in the T2I latent space.
\cref{fig:teaser:shift} illustrates this shift by projecting the T2I latents onto the 2D subspace found using supervised principal component analysis (SPCA) \cite{barshan2011supervised}.

\textbf{Finding the steering subspaces} does not require access to the weights of the target SID models but only their predicted hard labels, thereby making a black-box attack feasible \textbf{(addresses L1)}.
We compute these subspaces just once over a set of generated images and their corresponding SID predictions and universally apply them at inference time, thus bypassing the overhead of computing image-specific directions \textbf{(addresses L2)}. 
The universal applicability of \methodName{} also implies that the steered images and the set of direction-finding images can be heterogeneous. For example, the directions found from a set including animal and common object images can steer human images (see \cref{fig:teaser:pj}). In addition, we find that \methodName{} directions at lower resolutions remain valid even at higher resolutions, avoiding the costs associated with 1) generating the direction-finding image set in a higher resolution, and 2) finding the subspaces in higher-dimensional space.

\begin{tcolorbox}[width=\textwidth,
                  colback=skyblue,
                  colframe=skyblue,
                  colback=fillcolor7!30!green!20!white,
                  colframe=fillcolor7!30!green!20!white,
                  boxsep=5pt,
                  left=0pt,
                  right=0pt,
                  top=2pt]%
\noindent\textbf{Contributions.} 
\begin{itemize}[leftmargin=0.4cm]
    \setlength\itemsep{0pt}
    \item 
    We introduce \methodName{}, the first \textbf{(a) black-box} and \textbf{(b) distribution-based} unrestricted adversarial attack on synthetic image detectors, which enables computationally efficient \Rt{} on commercial and proprietary models, even with API-only access.
    \item \methodName{} \textbf{significantly enhances} the success rate of a T2I attacker in deceiving a target SID model, even when it is specifically tuned on images from the very same T2I model.
    \item \methodName{} enables \textbf{efficient attacks} for high-resolution images as \methodName{} directions are \textbf{transferable} across image resolutions.
    \item We achieve the \textbf{underlying objective of \Rt{}}, by using \methodName{}-steered samples to reduce the false negative rate of SID models. The images generated for the development of \methodName{} are to be released for the benefit of future research.
\end{itemize}
\end{tcolorbox}

\section{The Universal Unrestricted Attack on Synthetic Image Detector Problem}

\textbf{Notation.}\quad 
Scalars are denoted by lowercase letters (e.g., $\lambda$, $\tau$). Deterministic vectors and matrices are represented by boldface lowercase and uppercase letters, respectively (e.g., $\bm{x}$, $\bm{z}$ for vectors, and $\bm{H}$, $\bm{\Theta}$ for matrices). 
The element in row $i$ and column $j$ of matrix $\bm{M}$ is denoted by either $(\bm{M})_{ij}$ or $m_{ij}$. We use $\bm{I}_n$ (or $\bm{I}$) to denote the $n \times n$ identity matrix. Likewise, $\bm{1}_n$ and $\bm{0}_n$ represent $n\times 1$ vectors of ones and zeros, respectively. For a square matrix $\bm{K}$, the trace operator is denoted as $\Tr{\bm{K}}$. Sets and vector spaces are represented using calligraphic letters (e.g., $\mathcal{X}$, $\mathcal{Z}$).

\textbf{Problem Formulation.}\quad 
Let $\mathcal{C}$ denote a set of captions obtained from the distribution of textual descriptions $p(\bm c)$, serving as inputs to a T2I generative model $G$. Given a latent vector $\bm{z} \in \mathcal{Z}$ from the distribution of latents $p(\bm z)$ and a textual prompt $\bm c \in \mathcal{C}$, the T2I model generates an image following
$
G: \mathcal{Z}\times \mathcal{C}\rightarrow \mathcal{X},
$
where $\mathcal{X}$ is the space of images. 

For obtaining $P(\text{fake} \mid \bm x)$ from the perspective of an SID\footnote{We use SID to abbreviate both Synthetic Image Detection and Synthetic Image Detector.}, we assume access to a black-box detector model defined as
$
f:\mathcal{X}\rightarrow [0,1],
$
where, given an image $\bm x\in\mathcal{X}$ and a classification threshold $\tau\in(0,1)$, the SID labels $\bm x$ as \emph{real} if $f( \bm x ) < \tau$, and \emph{fake} or \emph{synthetic} otherwise. 

Let there be a mapping function $h$ within the T2I latent space parameterized by $\bm \Theta$, defined as 
\begin{equation}
    h_{\bm\Theta}\colon \mathcal{Z}\to\mathcal{Z}.
\end{equation}
The primary goal of our UA attack is to find the parameters of the mapping function $h_{\bm \Theta}$ such that latents mapped with $h_{\bm \Theta}$ generate images that are consistently misclassified by the SID model. This transformation of latents within the latent space is typically described as steering.
Formally, we frame the attack as the following optimization problem:

\begin{tcolorbox}[left=0pt,right=0pt,top=-7pt,bottom=2pt,colback=fillcolor7!30!green!20!white,colframe=fillcolor7!70!black,title=\begin{problem}\label{eq:redteam_obj}\emph{Unrestricted Adversarial Attack}\end{problem}]
\begin{align}
\max_{\bm{\Theta}}\; &\mathbb{E}_{\bm{z}\sim p(\bm{z}),\,\bm c\sim p(\bm c)}\Bigl[\mathbbm{1}\{f(G(h_{\bm \Theta}(\bm{z}), \bm c)) < \tau\}\Bigr],\nonumber
\end{align}
\end{tcolorbox}

where $\mathbbm{1}\{\cdot\}$ denotes the indicator function, returning $1$ when its argument is true and $0$ otherwise. 

A practical solution to \cref{eq:redteam_obj} requires two underlying constraints: 
\textbf{(C1) Black-box SID constraint:} the SID model must provide thresholded predictions over $f(\bm x)$ only, without any access to internal information (e.g., gradients, model weights), reflecting the settings of state-of-the-art proprietary SID models, and \textbf{(C2) Universal mapping constraint:} the mapping function $h_{\bm \Theta}$ must generalize across multiple images and prompts, avoiding gradient-based image-specific optimization that are computationally expensive.
Solving \cref{eq:redteam_obj} thus uncovers systematic vulnerabilities in SID models, enabling the creation of challenging adversarial examples to enhance their robustness.

\section{\methodName{}: An Approach to the Universal Unrestricted Attack Problem \label{sec:app}}
\begin{figure}
    \centering
    \vspace{-1em}
    \includegraphics[width=\linewidth]{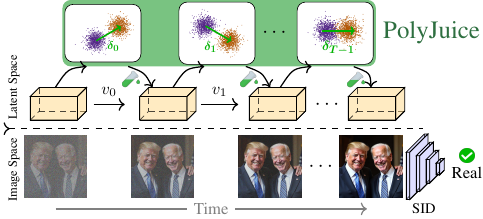}
    \caption{\textbf{Overview of \methodName{}:}  At each inference step $t$, \methodName{} manipulates the T2I latent using pre-computed direction $\bm \delta_t$ between predicted real and fake in order to deceive the target SID.}
    \vspace{-1em}
    \label{fig:overview}
\end{figure}
\textbf{Inapplicable Steering Approaches.}\quad
T2I models can be steered using their learned text conditions to modify certain attributes, such as age, hair color, style, etc. \cite{friedrich2023fair,friedrich2024auditing,zhang2024attention}. However, these methods are inapplicable for an abstract property like realness, which is unlikely to be explicitly introduced in the T2I training data.
Prior methods perform steering for novel concepts by relying on gradients from an external classifier (i.e., classifier guidance \cite{dhariwal2021diffusion}). This approach can be extended to deceive SID models by using gradients obtained from the target detector during the generation process. However, computing gradients requires white-box access to the SID, violating the black-box constraint \textbf{(C1)}.

\textbf{Black-Box Alternative.}\quad
Even though the T2I model has not learned realness as a semantic property to condition on, we find that the change in SID-predicted realness can be clearly identified as a distribution shift in the T2I model's latent space, as shown in \Cref{fig:teaser:shift}.
This shift can be identified using only hard labels from the target SID model, bypassing the need for white-box access \textbf{(satisfies C1)}. In contrast to image-specific mappings, the identified shift direction is universally applicable to arbitrary images \textbf{(satisfies C2)}. Our observations hint at the existence of a potential black-box method for performing universal attacks on SID models, consequently motivating \methodName{}. 

\subsection{Discovering the Universal Realness Shifts in Text-to-Image Latent Space}
To discover the universal shift, \methodName{} finds the subspace in the T2I latent space that has maximum statistical dependence on the label change of samples. Let $\bm Y = [\bm y^{(1)}, \bm y^{(2)}, \cdots, \bm y^{(n)}]^\top$ be the matrix of labels corresponding to the matrix of $n$ latents $\bm Z = [\bm z^{(1)}, \bm z^{(2)}, \cdots, \bm z^{(n)}]^\top$ where $\bm y^{(i)} \in \mathbb{R}^{d_Y}$ and $d_Y$ is the dimension of label space. 
There exists an orthogonal matrix $\bm U$ that transforms $\bm Z$ into a subspace where the variance of the labeled data is maximized, as determined by supervised principal component analysis (SPCA) \cite{barshan2011supervised}.
SPCA captures the dependence between the mapped samples, i.e., $\bm U^\top \bm Z$, and the corresponding labels $\bm Y$, using Hilbert-Schmidt Independence Criterion (HSIC) \cite{gretton2005measuring} as the dependence metric. The empirical version of HSIC is defined as
\begin{equation}
    \text{HSIC} =  \Tr{\bm H \bm K_{\bm{ZZ}} \bm H \bm K_{\bm{YY}}},
\end{equation}
where $\bm H = \bm I_n - \frac{1}{n}\ \bm 1_n \bm 1_n^\top$ is a centering matrix, $\bm K_{\bm{ZZ}}$ is a kernel matrix of the mapped data that is formulated as $\bm K_{\bm{ZZ}} = \bm Z^\top \bm U \bm U^\top \bm Z$  in case of a linear kernel, and $\bm K_{\bm{YY}} =  \bm Y^\top \bm Y$ is a kernel matrix of labels. Therefore, $\bm U$ can be calculated as
\begin{align}
\label{eq:U}
    \underset{\bm U}{\argmax}&\  \Tr{\bm U^\top \highll{$\bm Z \bm H \bm K_{\bm{YY}} \bm H \bm Z^\top$} \bm U }, \nonumber \\
    & \text{s.t.} \quad \bm U^\top \bm U = \bm I.
\end{align}

\textbf{The optimal solution for $\bm U$} can be obtained in closed-form by finding the eigenvectors corresponding to the $d$-largest eigenvalues of $\bm A:=$\highll{$\bm Z \bm H \bm K_{\bm{YY}} \bm H \bm Z^\top$}~\cite{lutkepohl1997handbook}, where $d$ is the dimensionality of the subspace. 
Here, since we only need a direction vector, we take a convex combination of the $d$ eigenvectors weighted by their corresponding eigenvalues $\{\sigma_k\}_{k=0}^{d-1}$ as
\begin{align}
   \bm \delta = \sum_{k=0}^{d-1} \sigma_k \ \bm U_k.
    \label{eq:delta}
\end{align}

\textbf{In case of time-varying T2I models} such as diffusion and flow-matching models,
the latent space is a time-indexed collection $\{\mathcal Z_t \}_{t=0}^{T-1}$, that we formulate as a single latent space $\mathcal Z = \oplus_{t=0}^{T-1} \mathcal Z_{t}$.
From a clean sample $\bm z_{T} \in \mathcal Z_T$, we can compute $\bm z_t \in \mathcal{Z}_t$ for any timestep $t$ as 
$
    \bm z_{t} = a_t \bm z_{T} +  b_t \bm \epsilon,
$
where $\bm \eps \sim \mathcal{N}(\bm 0, \bm I)$ and $(a_t, b_t)$ constitute the variance schedule of the T2I model. 
By computing a steering direction $\bm \delta_t$ as defined in \cref{eq:delta} within each $\mathcal Z_t$, we obtain \textbf{a set of steering directions} 
\begin{align} \label{eq:total delta}
\Delta = \{\bm \delta_{0}, \bm \delta_{1}, \ldots, \bm \delta_{T-1} \}.
\end{align}

\textbf{To solve \cref{eq:redteam_obj}}, we use the steering directions $\bm \delta_t \in \Delta$ to shift the latents in the direction of maximal change between true-positive (TP) and false-negative (FN) samples within each $\mathcal Z_t$. 
Given a set of generated images and predicted labels (0: real, 1: fake), for each timestep $t$, we solve \cref{eq:U} and use \cref{eq:delta} to compute $\bm \delta_t$. 
Therefore, the mapping function in \cref{eq:redteam_obj} becomes a set of time-varying mappings $ h_{\bm \Theta} = \{ h_{\bm\delta_t} \}_{t = 1}^{T-1}$, with $\bm \Theta = \Delta$ and each $h_{\bm\delta_t}$ given by
\begin{align}
h_{\bm\delta_t}(\bm z'_t) &= \bm z'_t + \hlight{$\lambda_t \bm \delta_t$},\quad t=1, \ldots, T-1
\label{eq:h}
\end{align}
where $\bm z'_t = h_{\bm \delta_{t-1}}(\bm z'_{t-1}) + \bm v_t$ given the T2I estimated velocity $\bm v_t$, and $\lambda_t \in [0, \infty)$ controls the steering strength at each time step.
As shown in \cref{fig:overview}, at each sampling step, \methodName{} uses $\bm \delta_t$ to steer the latent originally updated by $\bm v_t$, to find an attack image that is misclassified as real by the target SID model.
\textbf{These attack samples help to improve the detector} by finding its semantic and non-semantic failure modes, as corroborated by our results in \cref{sec:res:tradeoff}.

\subsection{Transferability of \methodName{} Directions from Low to High Image Resolution\label{sec:app:res}}
Calculating latent directions separately for each desired image resolution incurs computational costs due to the requirement of creating new datasets, adding the noise corresponding to each timestep, and recomputing steering directions. To bypass these costs, we propose \emph{resolution transferability}, which involves computing steering directions once at a base resolution (e.g., $256$$\times$$256$), followed by upscaling them using interpolation and applying them to higher resolution images (e.g., $1024$$\times$$1024$).

Formally, let $\bm \delta_t \in \mathbb{R}^{C \times H \times W}$ represent the steering direction vector at timestep $t$, calculated at a base resolution $H \times W$ with channel dimension $C$. When applying this direction at a higher target resolution $H' \times W'$, we first upscale the direction via interpolation as follows:
\begin{equation}
\bm \delta'_t = \text{Interp}(\bm \delta_t; H', W'),
\label{eq:interp}
\end{equation}
where $\text{Interp}(\cdot)$ denotes spatial interpolation. The upscaled directions $\bm \delta'_t \in \mathbb{R}^{C \times H' \times W'}$ are then applied to the corresponding latents at the higher resolution.

The key insight behind resolution transferability stems from the fact that the KL-regularized autoencoders \citep{rombach2022high} used by recent T2I models focus on perceptual compression and are thus resolution-invariant. T2I latents across different resolutions are likely to maintain similar spatial properties and semantic structures, which makes the computed direction vectors transferable between resolutions to some extent. By employing resolution transferability, \methodName{} avoids the significant computational overhead associated with the generation and preprocessing of a high-resolution synthetic image dataset, i.e., steps that are needed before finding attack directions.

\section{Experimental Evaluation \label{sec:res}}
We evaluate the effectiveness of \methodName{} across varying combinations of T2I and SID models. For T2I generation, we utilize \sd{} \citep{esser2024scaling}, \dev{} \citep{flux}, and \sch{} \citep{flux}, at three image resolutions, $256$$\times$$256$, $512$$\times$$512$, and $1024$$\times$$1024$, respectively. As the target synthetic image detector, we employ Universal Fake Detector (UFD) \citep{ojha2023towards} and RINE \citep{koutlis2024leveraging}---two recent SID methods trained on common objects. For real images and text prompts, we use the Common Objects in Context (COCO) dataset \citep{lin2014microsoft}, which contains captioned images of common objects.
Although there are alternative T2I datasets (e.g., LAION 5B \citep{laion}), these datasets contain significant amounts of digitally created content.

\noindent\textbf{Experiment Setting.}\quad 
We strictly assume black-box access to both UFD and RINE; the weights and gradients remain hidden, and querying the SIDs with an image only returns a hard label $Y = 1$ (fake) or $Y = 0$ (real). Each detector predicts a confidence score and then uses a threshold $\tau$ to provide the label $Y$. For RINE, we set this threshold to 0.5, following the same settings used by \citet{koutlis2024leveraging}. For UFD, we follow the calibration approach of \citet{ojha2023towards} to adjust the threshold using real images from the COCO training set and a mixed set of generated images from all three T2I models.
For each pair of T2I generator and SID model, the set of directions $\Delta$ is calculated from 20K true-positive (TP) and 20K false-negative (FN) examples generated from COCO training captions. 
For evaluation, we generate 1000 images from text descriptions of the COCO validation set; these captions are also used to perform the attack using \methodName{}.
As our primary \textbf{evaluation metric}, we adopt \emph{success rate} (i.e. \emph{false negative rate}), which is defined as the proportion of fake images the SID erroneously detects as real.

\begin{figure}
    \centering
    \vspace{-2em}
    \includegraphics[width=0.9\linewidth]{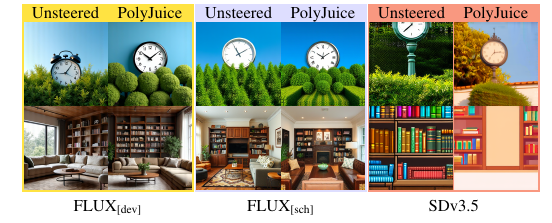}
    \caption{\textbf{Unsteered vs. \methodName{}-steered T2I attacks} against UFD. Although UFD detects the unsteered generated images as fake, \methodName{}-disguised samples successfully deceive the detector.}
    \label{fig:qual}
\end{figure}

\subsection{How Successful is \methodName{} in Attacking Synthetic Image Detectors? \label{sec:res:main}}

\renewcommand{\arraystretch}{1.0}
\begin{table}[htbp]
  \centering
  \vspace{-1em}
  \caption{Comparison of \methodName{} and baselines w.r.t. success rate (\%) on COCO validation set.
  \label{tab:main-results}}
    \resizebox{0.8\columnwidth}{!}{%
    \begin{tabular}{l c c cc c cc}
    \toprule
     & \multirow{2}[3]{*}{T2I Model} && \multicolumn{2}{c}{UFD \cite{ojha2023towards}} && \multicolumn{2}{c}{RINE~\cite{koutlis2024leveraging}} \\
     \cmidrule{4-5} \cmidrule{7-8}
     & && No Steering & \methodName{} (ours) && No Steering & \methodName{} (ours) \\
\midrule
    \multirow{3}{*}{\begin{sideways}\footnotesize \scalebox{0.9}{256$\times$256}\end{sideways}} 
     & \sd{} && 12.8 & 80.6 (\textcolor{red!50!black}{\textbf{+67}}) && 15.3 & 99.7 (\textcolor{red!50!black}{\textbf{+84}})\\
     & \dev{} && 67.6  &  96.3 (\textcolor{red!50!black}{\textbf{+28}}) && 52.4 & 81.2 (\textcolor{red!50!black}{\textbf{+28}})\\
     & \sch{} && 61.7 &  83.4 (\textcolor{red!50!black}{\textbf{+21}}) && 45.4 & 73.8 (\textcolor{red!50!black}{\textbf{+28}})\\
     \midrule
     \multirow{3}{*}{\begin{sideways}\footnotesize \scalebox{0.9}{512$\times$512}\end{sideways}} 
     & \sd{} &&  30.5  &  85.0 (\textcolor{red!50!black}{\textbf{+54}}) &&  26.7 & 99.6 (\textcolor{red!50!black}{\textbf{+72}})\\
     & \dev{} &&  84.0   & 98.9 (\textcolor{red!50!black}{\textbf{+14}})  &&  77.2 & 96.7 (\textcolor{red!50!black}{\textbf{+19}}) \\
     & \sch{} &&  72.7 & 85.1 (\textcolor{red!50!black}{\textbf{+12}})  &&  62.9 & 84.1  (\textcolor{red!50!black}{\textbf{+21}})\\
     \midrule
     \multirow{3}{*}{\begin{sideways}\footnotesize \scalebox{0.8}{1024$\times$1024}\end{sideways}} 
     & \sd{}  &&  59.3   &  93.3  (\textcolor{red!50!black}{\textbf{+34}})   &&  51.0   &  99.8   (\textcolor{red!50!black}{\textbf{+48}})\\
     & \dev{} &&  75.6   &  98.4  (\textcolor{red!50!black}{\textbf{+22}})   &&  82.4   &  94.9   (\textcolor{red!50!black}{\textbf{+12}}) \\
     & \sch{} &&  72.1   &  84.0  (\textcolor{red!50!black}{\textbf{+11}})   &&  69.8   &  95.6   (\textcolor{red!50!black}{\textbf{+25}})\\
     \midrule
     \multicolumn{2}{c}{\ AVG \textcolor{gray}{$\pm$ STD}} &&  59.6 \textcolor{gray}{$\pm$ 21.8}  & 89.4 \textcolor{gray}{$\pm$ 6.8}  &&  53.7 \textcolor{gray}{$\pm$ 21.1} & 91.7 \textcolor{gray}{$\pm$ 9.0}\\
\bottomrule       
    \end{tabular}
    }
    \vspace{-1.5em}
\end{table}

\Cref{tab:main-results} compares \methodName{}-steered T2I models against unsteered baselines in terms of their ability to deceive SID models.
We observe that, among the three unsteered T2I generators, \sd{} has the lowest success rate---suggesting that UFD and RINE are quite effective at detecting \sd{}-generated fakes, while \dev{} and \sch{}-generated images are harder to detect. However, by steering the image generation process of each T2I model with \methodName{}, the success rate in deceiving the SID models is significantly boosted, even in the case of the easily-detected \sd{}. To perform attacks for $512$ and $1024$ resolutions, we compute $\Delta$ from $256$$\times$$256$ synthetic images, and transfer them through interpolation (\cref{eq:interp}).
Although RINE is more robust than UFD against \emph{unsteered} T2I models, we find that both UFD and RINE are similarly vulnerable against \methodName{}-steered attacks (see \cref{tab:main-results} last row for avg. $\pm$ std.). 
From the perspective of the T2I generators, \dev{} consistently outperforms \sch{}, potentially due to having more inference steps (50 vs. 4) to apply \methodName{}.  
\Cref{fig:qual} shows some qualitative examples corresponding to \Cref{tab:main-results}, where the unsteered images are detected as fake.
We observe that, regardless of whether the original image looks highly realistic or obviously synthetic, \methodName{} is able to fool the target detector, demonstrating its effectiveness. 

\remark{\methodName{} boosts attack success rate of T2I models against SIDs by up to \textcolor{red!50!black}{\textbf{84\%}}.}

\subsection{How Effective is \methodName{} When Applied on a T2I Model-Specific Detector? \label{sec:res:ft}}

\begin{wraptable}{r}{0.55\linewidth}
\vspace{-1.5em}
\caption{\footnotesize \methodName{} against T2I-specific SIDs at $256$$\times$$256$.\label{tab:ft-res}}
\vspace{-0.5em}
\resizebox{\linewidth}{!}{%
\begin{tabular}{lcccccc}
        \toprule
        \multirow{2}[3]{*}{Model} && \multicolumn{2}{c}{UFD~\cite{ojha2023towards}} && \multicolumn{2}{c}{RINE\cite{koutlis2024leveraging}} \\
        \cmidrule{3-4} \cmidrule{6-7}
         && No Steering & \methodName{} && No Steering & \methodName{} \\
        \midrule
        \sd{}     &&  18.3 & 90.8 (\textcolor{red!50!black}{\textbf{+72}}) && 8.0  & 64.8  (\textcolor{red!50!black}{\textbf{+52}})\\
        \dev{}     &&  33.7 & 82.1 (\textcolor{red!50!black}{\textbf{+48}}) &&  21.8 & 86.6 (\textcolor{red!50!black}{\textbf{+64}}) \\
        \sch{}     && 40.0  & 64.9 (\textcolor{red!50!black}{\textbf{+24}}) &&  20.5 &  29.3 (\textcolor{red!50!black}{\textbf{+8}})\\
        \bottomrule
        \end{tabular}%
        }
\vspace{-1.5em}
\end{wraptable}
\textbf{Motivation.}\quad We aim to investigate the extreme scenario where the target SID model is \emph{tuned} using images generated by the very same T2I model employed in the attack.

For each T2I model $G$, we first calibrate the thresholds for both UFD and RINE using a set of real images from COCO and a set of generated images from $G$.
The goal of the calibration is to improve the overall \emph{accuracy} of the detector in distinguishing real images from fake images generated by a specific T2I model.
Next, we recompute the set $\Delta$ by obtaining a new set of hard labels from the calibrated models. We then evaluate \methodName{} on the updated SID models.
\Cref{tab:ft-res} demonstrates \methodName{}'s performance against the tuned SID models. Compared to \cref{tab:main-results}, we observe that finding a new threshold significantly improves the detection performance in all cases except for UFD against \sd{}.
 Both UFD and RINE significantly improve their detection of fake images from \dev{} and \sch{} under the new settings. In addition, RINE demonstrates an overall stronger detection capability compared to UFD. Against the improved detectors, \methodName{} still notably boosts the success rate, as much as $72\%$ (in the case of \sd{}). In this extreme case, we find that \sch{} shows the least success among the three T2I models. 
 Our combined observations from \cref{tab:ft-res} and \cref{sec:res:main} suggest that \sch{} has limited steerability, potentially due to fewer inference steps.  

\remark{Even in the extreme case when the target SID model is calibrated on a specific T2I model $G$, \methodName{} can still improve the FNR up to 72\% while using $G$ for generating its attacks.}

\begin{figure}[t!]
    \centering
    \vspace{-3em}
    \includegraphics[width=0.99\linewidth]{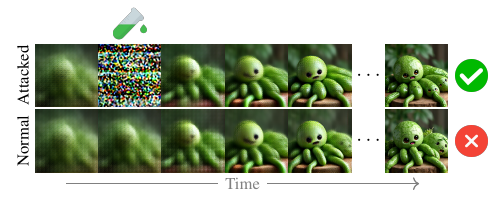}
    \caption{\textbf{Estimated clean images at various timesteps} for \dev{}, where bottom and top rows depict unsteered and \methodName{}-steered generation, respectively.
    \label{fig:steering-vis}}
    \vspace{-1.5em}
\end{figure}

\subsection{How Effective is \methodName{} in Reducing False Negative Rate of Existing SIDs?}
\label{sec:res:tradeoff}

\renewcommand{\arraystretch}{0.8}
\begin{wraptable}{r}{0.55\linewidth}
  \centering
  \vspace{-1.5em}
  \caption{Comparison of SID models' FNR before and after \methodName{}-aided calibration.
  \vspace{-0.5em}
  \label{tab:improve-sid}}
    \resizebox{\linewidth}{!}{%
    \begin{tabular}{l c c cc c cc}
    \toprule
     & \multirow{2}[3]{*}{T2I Model} && \multicolumn{2}{c}{UFD \cite{ojha2023towards}} && \multicolumn{2}{c}{RINE~\cite{koutlis2024leveraging}} \\
     \cmidrule{4-5} \cmidrule{7-8}
     & && Pre-Cal. & \includegraphics[width=0.8em]{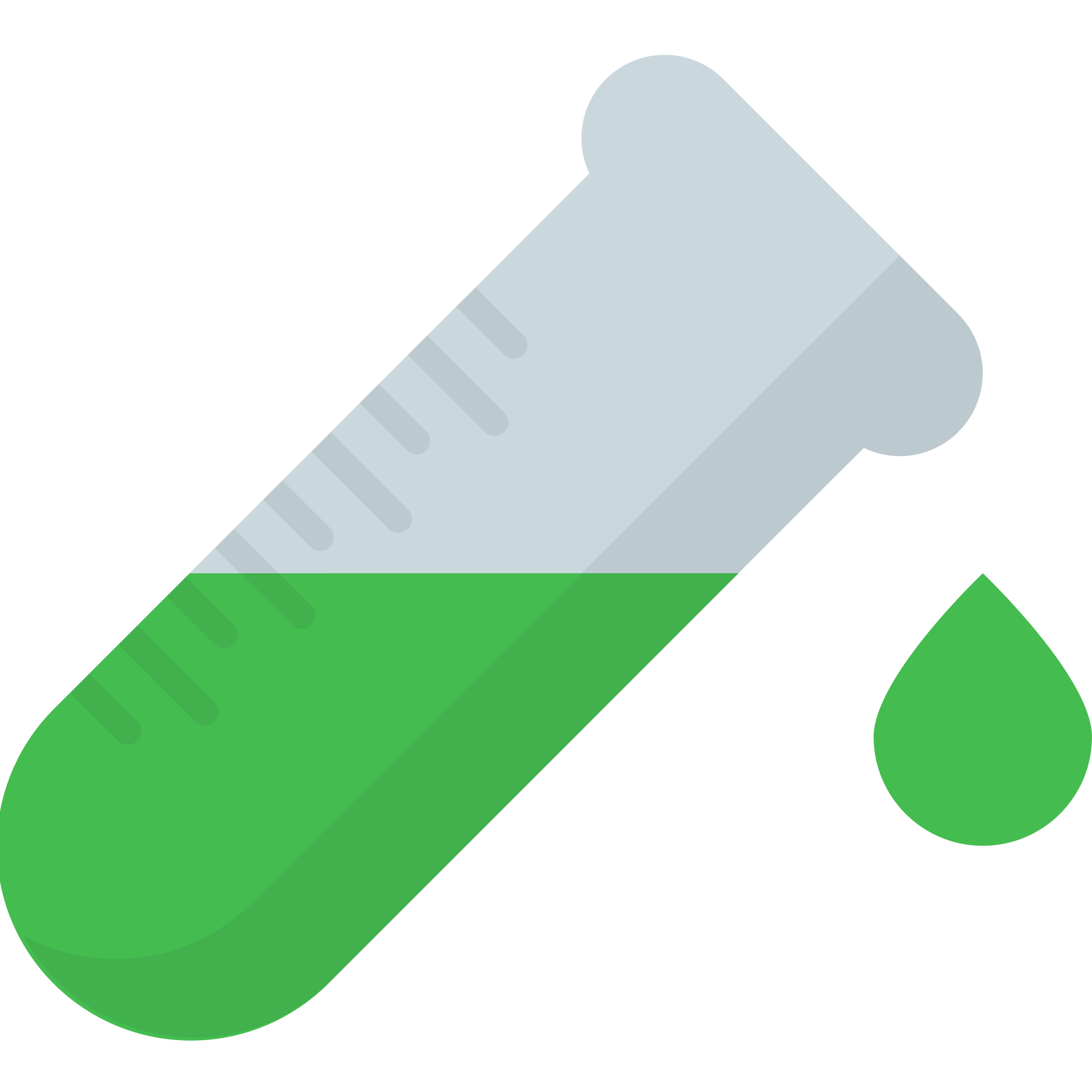}-Cal. && Pre-Cal. & \includegraphics[width=0.8em]{figs/pj.png}-Cal. \\
\midrule
    \multirow{3}{*}{\begin{sideways}\footnotesize 256 \end{sideways}} 
     & \sd{}  && 13.4 &  7.5 (\textcolor{fillcolor7!60!green!80!black}{\textbf{-5}}) && 15.1  & 3.8   (\textcolor{fillcolor7!60!green!80!black}{\textbf{-11}})\\
     & \dev{} && 69.2 & 47.0  (\textcolor{fillcolor7!60!green!80!black}{\textbf{-22}})&& 52.0  & 21.8  (\textcolor{fillcolor7!60!green!80!black}{\textbf{-30}})\\
     & \sch{} && 64.3 & 43.7  (\textcolor{fillcolor7!60!green!80!black}{\textbf{-20}}) &&  39.6 & 18.4 (\textcolor{fillcolor7!60!green!80!black}{\textbf{-21}})\\
     \midrule
     \multirow{3}{*}{\begin{sideways}\footnotesize 512 \end{sideways}} 
     & \sd{}  && 31.1 &  16.1 (\textcolor{fillcolor7!60!green!80!black}{\textbf{-15}}) &&  17.8 & 4.7  (\textcolor{fillcolor7!60!green!80!black}{\textbf{-13}})\\
     & \dev{} && 86.2 &  70.3 (\textcolor{fillcolor7!60!green!80!black}{\textbf{-15}}) &&  69.4 & 41.4 (\textcolor{fillcolor7!60!green!80!black}{\textbf{-28}}) \\
     & \sch{} && 71.8 &  55.2 (\textcolor{fillcolor7!60!green!80!black}{\textbf{-16}}) &&  50.9 & 25.4 (\textcolor{fillcolor7!60!green!80!black}{\textbf{-25}}) \\
\bottomrule
    \end{tabular}
    }
    \vspace{-1em}
\end{wraptable}

To observe the effectiveness of \methodName{} in improving target SID models, we first attack the target detectors with \methodName{}-steered T2I models.
Next, we calibrate each SID using a combination of (i) real images from COCO, (ii) regular synthetic images from a T2I model (i.e. unsteered T2I models), and (iii) \methodName{}-steered successful attack images.
In \cref{tab:improve-sid}, we compare the \emph{pre-} and \emph{post-}calibration SID models (Pre-Cal. and \includegraphics[width=0.8em]{figs/pj.png}-Cal.) on a set of synthetic COCO validation images at two different resolutions. We find that after \methodName{}-aided calibration, both UFD and RINE show significant improvements in detecting regular (unsteered) generated images.

\remark{Calibrating SID models using \methodName{} reduces their vulnerabilities by up to \textbf{30\%}.}

\subsection{How Transferable are the Directions from Lower to Higher Resolutions?\label{sec:res-transfer}}

\begin{wraptable}{r}{0.55\linewidth}
\vspace{-1em}
\caption{\footnotesize Resolution transferability of \methodName{} on $512 \times 512$.\label{tab:res-transfer}}
\vspace{-0.5em}
\resizebox{\linewidth}{!}{%
\begin{tabular}{lccccccc}
        \toprule
        \multirow{2}[3]{*}{Model} &&  \multicolumn{3}{c}{RINE\cite{koutlis2024leveraging}} \\
        \cmidrule{3-5}
         && No Steering & Original & Transferred \\
        \midrule
        \sd{}      && 26.7  & 77.6  & \textbf{99.6}  \\
        \dev{}     && 77.2  & 95.7  & \textbf{96.7} \\
        \sch{}     && 62.9  & 79.9  &  \textbf{84.1} \\
        \bottomrule
        \end{tabular}%
        }
\vspace{-0.3cm}
\end{wraptable}
We evaluate the \emph{transferability} (\cref{sec:app:res}) of low-resolution $256$$\times$$256$ directions by comparing their attack performance against attacks generated by $512$$\times$$512$ directions, as shown in \Cref{tab:res-transfer}. Both the transferred and original directions enable \methodName{} to significantly improve the FNR over the unsteered baseline. Further, \methodName{} attacks using transferred low-resolution directions achieve a comparable or higher FNR than attacks with original high-resolution directions. This result suggests that the error in SPCA-discovered directions is proportional to the dimensionality of its input space due to the curse of dimensionality \cite{koppen2000curse}.

\remark{\methodName{} directions transferred from lower resolutions achieve success rates comparable to or better than original high-resolution directions, while being less expensive to find.}

\subsection{How Does \methodName{} Modify the Image Generation Process?}

\begin{wrapfigure}{r}{0.45\columnwidth}
  \centering
  \vspace{-1.3em}
  \begin{minipage}[b]{0.49\linewidth}
    \includegraphics[width=\linewidth]{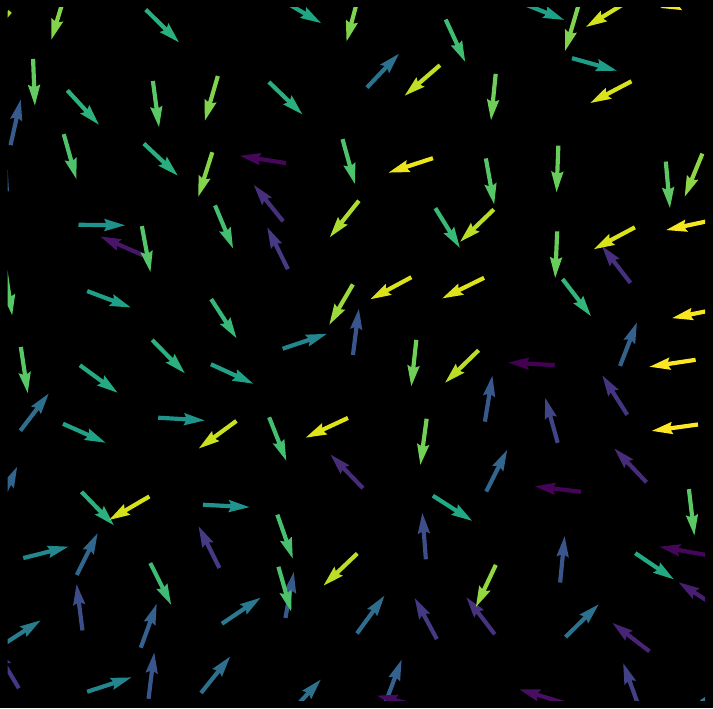}
  \end{minipage}
  \hfill
  \begin{minipage}[b]{0.49\linewidth}
    \includegraphics[width=\linewidth]{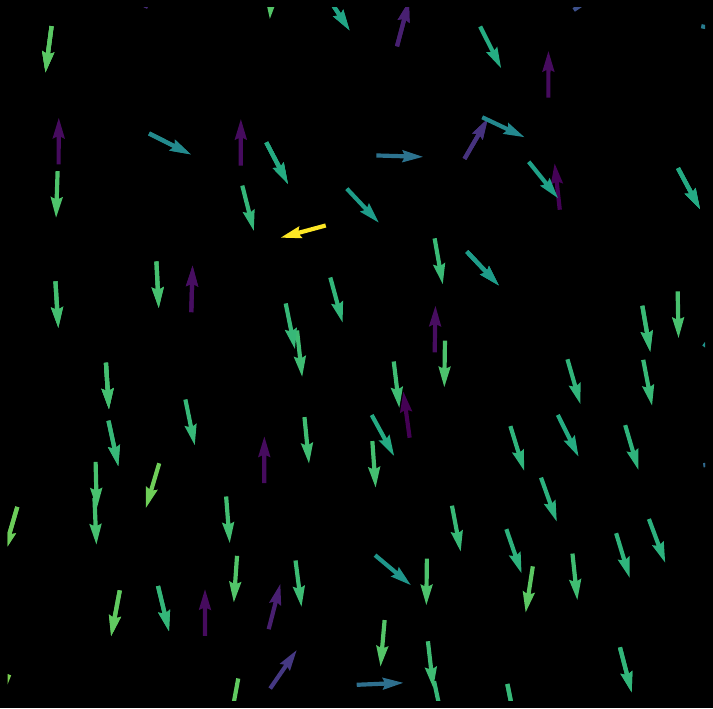}
  \end{minipage}
  \caption{\footnotesize (left) Unsteered (right) \methodName{}-steered samples projected on 2D subspace.\label{fig:vector}}
  \vspace{-1em}
\end{wrapfigure}
To better understand how \methodName{} works, in \cref{fig:steering-vis,fig:vector}, we illustrate the effect of the steering direction, $\bm \delta_t$, in the T2I image space and the projected latent space, respectively.
In \cref{fig:steering-vis}, we visualize the estimated clean images at different timesteps.  \methodName{} steers the latent towards the blind spot of the target SID (at the second image), resulting in a successful attack (top) that slightly differs from the unsteered counterpart (bottom) while maintaining the semantics of the image. 
In \cref{fig:vector}, we provide an SPCA-based 2D visualization of the update directions of the latents (colored by angle), where (left) and (right) depict unsteered and \methodName{}-steered, correspondingly.
We observe that a majority of \methodName{}-steered updates are aligned along a common direction---the direction that leads to deceiving the SID. See Appendix for more details.

\subsection{ How Does \methodName{} Affect the Spectral Fingerprint of the T2I Models?}

\begin{figure}[h]
    \centering
    \vspace{-0.5em}
    \begin{subfigure}[t]{0.24\linewidth}
        \includegraphics[width=\linewidth]{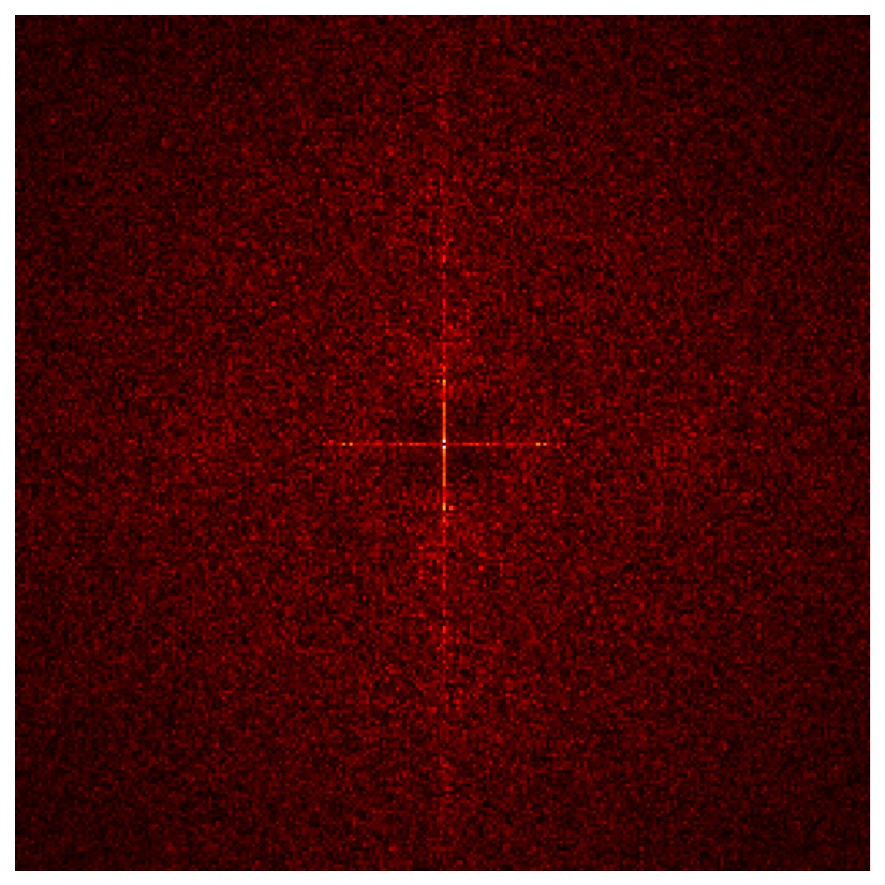}
        \caption{Real Samples\label{fig:freq:real}}
    \end{subfigure}
    \hfill
    \begin{subfigure}[t]{0.24\linewidth}
        \includegraphics[width=\linewidth]{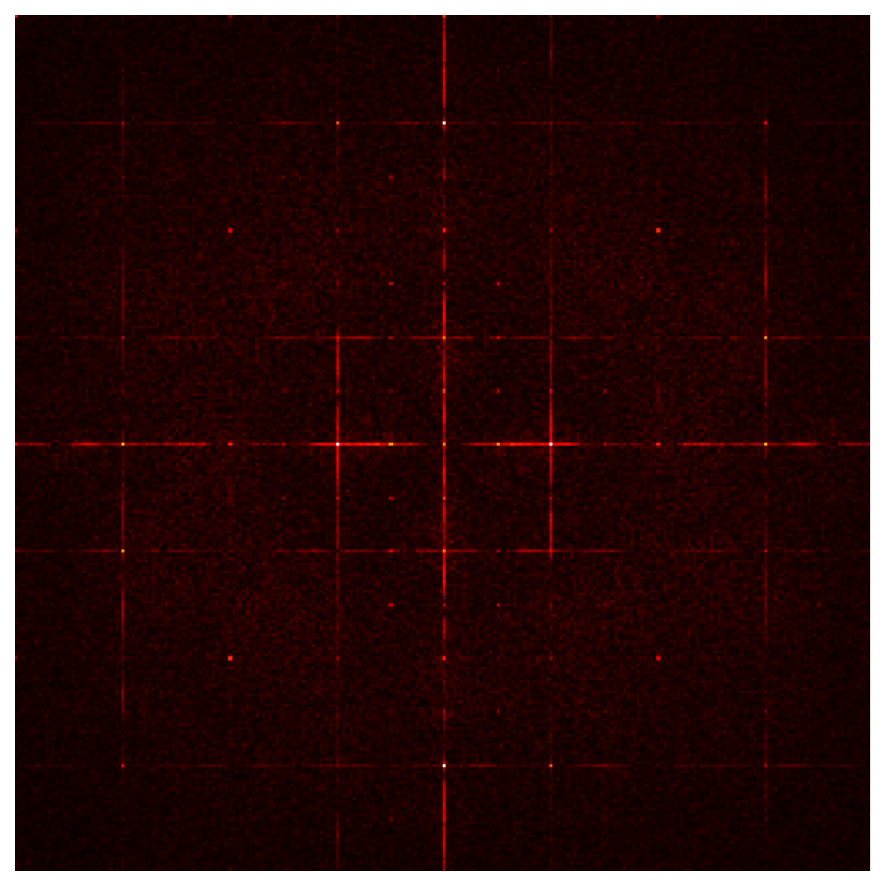}
        \caption{TP Samples\label{fig:freq:tp}}
    \end{subfigure}
    \hfill
    \begin{subfigure}[t]{0.24\linewidth}
        \includegraphics[width=\linewidth]{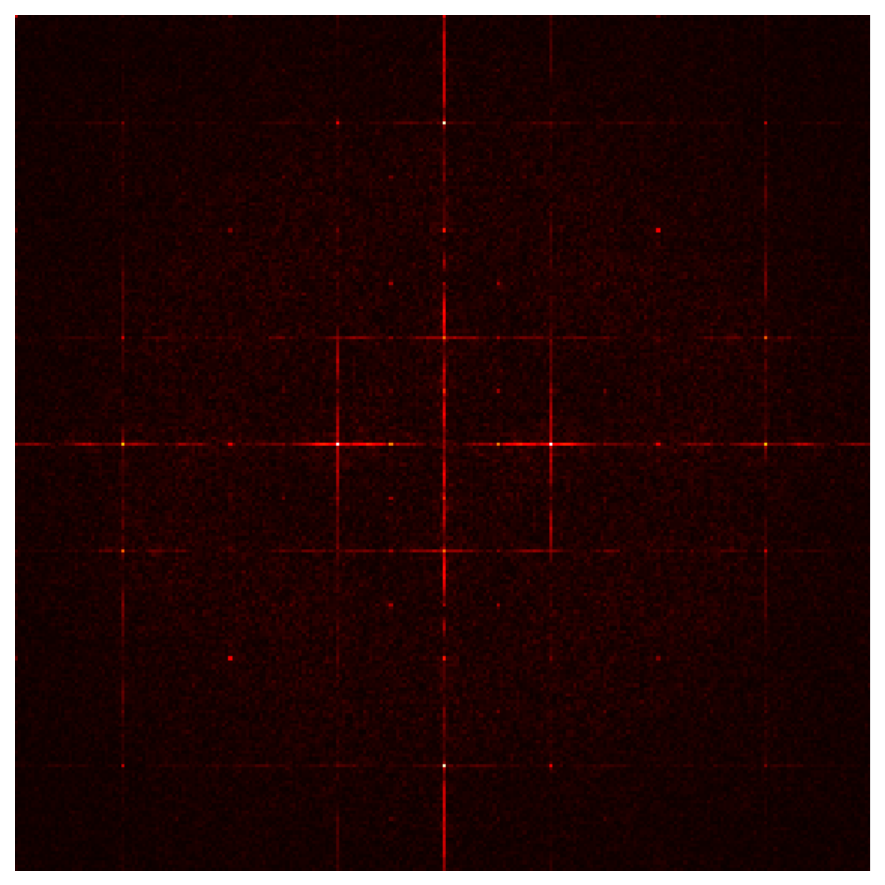}
        \caption{FN Samples\label{fig:freq:fn}}
    \end{subfigure}
    \hfill
    \begin{subfigure}[t]{0.24\linewidth}
        \includegraphics[width=\linewidth]{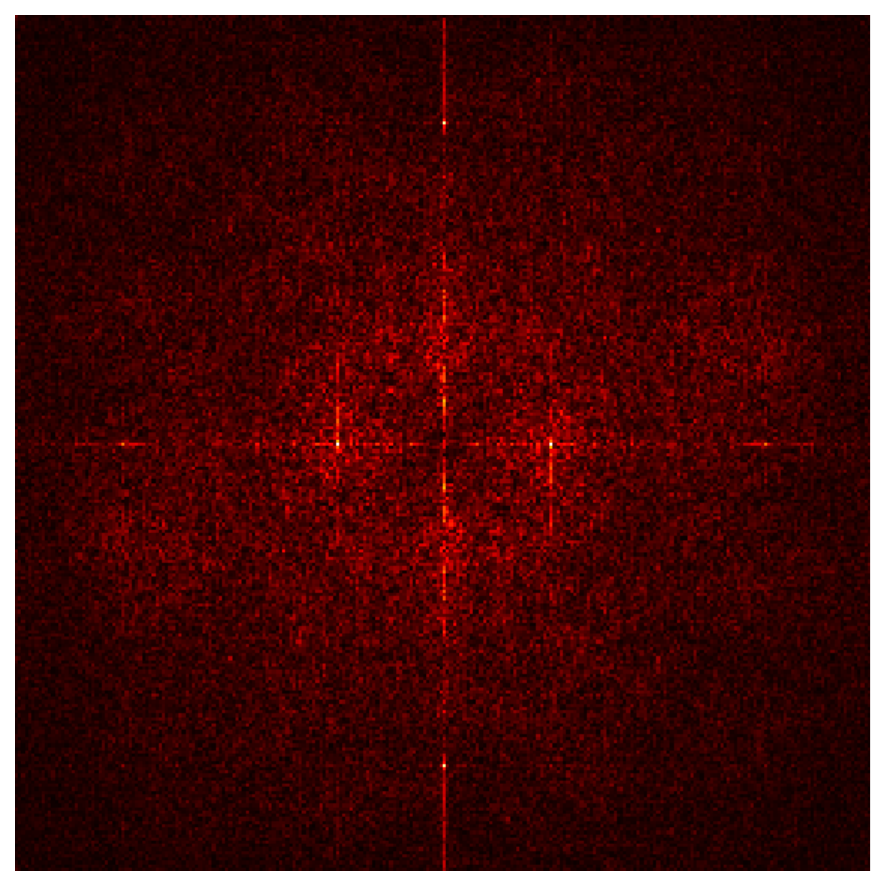}
        \caption{\methodName{} Attacks\label{fig:freq:att}}
    \end{subfigure}
    \caption{\textbf{Average Frequency Spectra} of COCO images and generated counterparts. }
    \label{fig:freq}
\end{figure}

Prior work \cite{ojha2023towards,corvi2023detection} uncovers the existence of \emph{fingerprints} left by T2I models through spectral analysis on the synthetic and real images. In \cref{fig:freq}, we present a similar approach on (a) the set of real images, (b) the \sd{}-generated images correctly detected by UFD (TP), (c) generated images misclassified by UFD (FN), and (d) successful \methodName{} attacks. First, we observe that there is a clearly noticeable difference between the real images and the unsteered \sd{}-generated images (TP + FN), which is the fingerprint of \sd{}. However, it can be seen that \methodName{}  obfuscates this fingerprint, as the residuals of \methodName{}-steered images (\cref{fig:freq:att}) appear closer to the real image spectrum (\cref{fig:freq:real}).  We provide implementation details and spectral analysis for other T2I models in the Appendix.

\remark{\methodName{} disguises the fingerprints left by T2I models in the frequency domain.}

\subsection{How Do \methodName{} Attacks Appear to the Eyes of the Target SID?}
\begin{wrapfigure}{r}{0.33\columnwidth}
    \centering
    \vspace{-1.3em}
    \includegraphics[width=\linewidth]{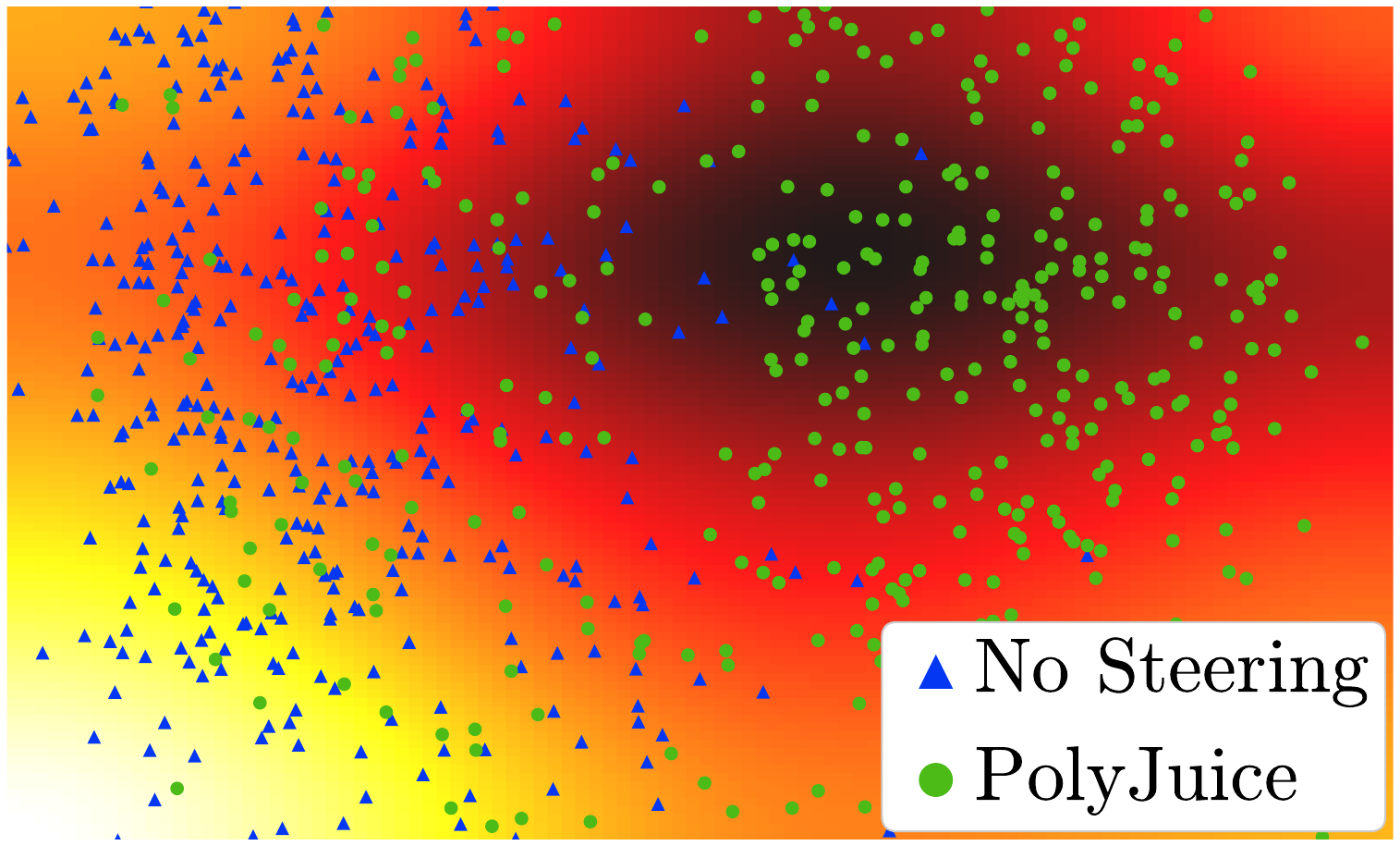}
    \caption{\footnotesize CLIP's embedding space.}
    \label{fig:clip-vis}
    \vspace{-2em}
\end{wrapfigure}
In \cref{fig:clip-vis}, we provide a 2D visualization of \methodName{}-steered attacks and unsteered samples from \sd{} in the embedding space of CLIP ViT-L/14~\cite{radford2021learning} using SPCA dimensionality reduction. We illustrate the prediction landscape of the target SID using a kernel density estimate, where darker areas denote input regions with high density of predicted realness. From \cref{fig:clip-vis}, we observe that there is a region of perceived realness in the SID's input space that remains unexplored by the unsteered T2I model. \methodName{} is able to \emph{identify} this region and \emph{exploit} it to deceive the SID.

\remark{\methodName{} steers samples into failure regions missed by standard T2I generation.}

\begin{figure}[h!]
\vspace{-3em}
    \centering
    \includegraphics[width=0.85\linewidth]{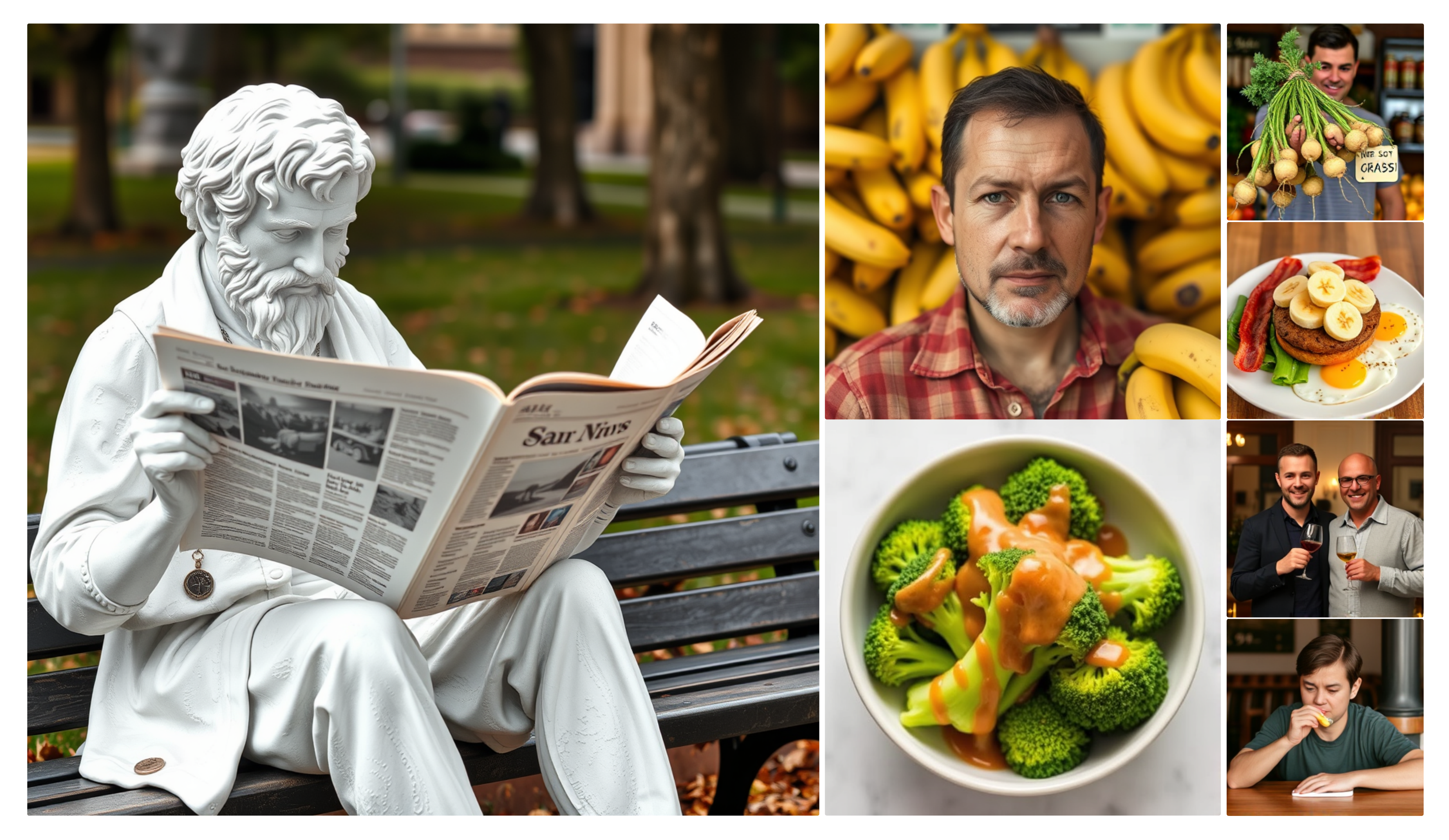}
    \caption{\small Attacks generated by \methodName{}-steered \sch{} model that were able to deceive RINE.}
    \label{fig:sch-rine-qua1}
    \vspace{-1em}
\end{figure}

\section{Does \methodName{} Conserve Image Quality? \label{app:reb:image-quality}}
As the goal of red teaming is to discover and mitigate the failure modes, finding attacks that are unrealistic (e.g., cartoonish features that fool the SID) is also important, since it reveals the failure mode of the target SID. An ideal SID must trivially detect these unrealistic images as fake. However, it is still important to generate attacks that are \emph{prompt faithful}, and of \emph{reasonable image quality}.

\begin{table}[h!]
\vspace{-1em}
\centering
\small %
\caption{Image quality comparison for images generated by unsteered \sch{} and \methodName{}, evaluated using various distribution-based metrics.}
\begin{tabular}{l c c c c c c}
\toprule
\textbf{Method} & \textbf{FID}~$\downarrow$& \textbf{cFID}~$\downarrow$& \textbf{Prec.}~$\uparrow$ & \textbf{Rec.}~$\uparrow$ & \textbf{Den.}~$\uparrow$ & \textbf{Cvg.}~$\uparrow$ \\
\midrule
Unsteered & 17.65  & 17.81 & 0.495 & 0.485 & 0.585 & 0.801 \\
PolyJuice & 17.23 & 17.41 & 0.485 & 0.498 & 0.764 & 0.794 \\
\bottomrule
\end{tabular}
\label{tab:fid}
\end{table}

\noindent\textbf{Distribution-level image quality.}\quad We compare an unsteered \sch{} and a \methodName{}-steered \sch{} against the RINE detector, at $256 \times 256$ image resolution. We use a randomly sampled set of 50K real images from the COCO dataset, and the same number of images from unsteered and \methodName{}-steered \sch{} respectively, and present our findings in \cref{tab:fid}. First, we compute both classical FID \citep{fid} and CleanFID \citep{cleanfid}, and find a similar trend: the FID w.r.t. real COCO images overall remains quite similar between both methods, suggesting similar quality (also validated by similar Precision (Prec.) and Recall (Rec.) scores \cite{ipr}).
These metrics suggest that \textbf{\methodName{} conserves image quality}, which rules out the possibility of \methodName{} deceiving SIDs by causing image degradations. Further, when comparing density and coverage \citep{dencov}, we notice that \methodName{} has a negligible effect on Coverage (Cov.), but yields a significantly improved Density (Den.) score.
This suggests that \methodName{}-steered images are more likely to reside in neighborhoods that are densely packed with real data points---which is aligned with the goals of \Rt{}. We show a few qualitative examples of attacks generated with \methodName{}-steered \sch{} in \cref{fig:sch-rine-qua1}; more examples are shown in \cref{app:qual}.

\begin{table}[!h]
  \vspace{-1em}
  \centering
  \caption{\small
  Image-level quality and prompt faithfulness assessment of \methodName{} vs. unsteered \dev{} using CLIP-based scores.}
  \label{tab:r2}
\resizebox{\columnwidth}{!}{%
  \begin{tabular}{l*{12}{c}}
    \toprule
    & \multicolumn{6}{c}{UFD} & \multicolumn{6}{c}{RINE} \\
    \cmidrule(lr){2-7}\cmidrule(lr){8-13}
    & \multicolumn{2}{c}{SR} & \multicolumn{2}{c}{CLIP-IQA} & \multicolumn{2}{c}{CLIP Score}
    & \multicolumn{2}{c}{SR} & \multicolumn{2}{c}{CLIP-IQA} & \multicolumn{2}{c}{CLIP Score} \\
    \cmidrule(lr){2-3}\cmidrule(lr){4-5}\cmidrule(lr){6-7}
    \cmidrule(lr){8-9}\cmidrule(lr){10-11}\cmidrule(lr){12-13}
    Res & Unst. & PolyJ. & Unst. & PolyJ. & Unst. & PolyJ. & Unst. & PolyJ. & Unst. & PolyJ. & Unst. & PolyJ. \\
    \midrule
    256  & 67.6 & {96.3} & 0.8427 & {0.8410} & 30.57 & {30.45}
         & 52.4 & {81.2} & 0.8427 & {0.8457} & 30.57 & {30.52} \\
    512  & 84.0 & {98.9} & 0.8535 & {0.8487} & 30.92 & {30.94}
         & 77.2 & {96.7} & 0.8535 & {0.8526} & 30.92 & {30.84} \\
    1024 & 75.6 & {98.4} & 0.8657 & {0.8633} & 30.86 & {30.91}
         & 82.4 & {94.9} & 0.8657 & {0.8503} & 30.86 & {30.69} \\
    \bottomrule
  \end{tabular}
  }%
\end{table}

\noindent\textbf{Image-level quality and prompt faithfulness.}  
 To measure per-image quality, we use the \textbf{CLIP IQA score} \citep{wang2023exploring}; for evaluating prompt faithfulness, we compute the \textbf{CLIP alignment score} between the generated attacks and the input prompts and present the results in \cref{tab:r2}. We find that the CLIP scores and CLIP-IQA scores are comparable to those of the unsteered models, suggesting that PolyJuice is not trivially evading detection by degrading the images.

\section{Related Work}
\textbf{Finding Directions from Images.}\quad 
Some prior work \cite{jain2022distilling,zhang2023iti} aim to identify latent space directions corresponding to specific visual attributes.
Similar to \methodName{}, \citet{zhang2023iti} also find directions from a set of images. But unlike \methodName{}, these directions are found in the CLIP's embedding space, with the motivation of addressing biases in T2I generation.
More similar to our application, \citet{jain2022distilling} use linear SVM to find a hyperplane that best separates correct from incorrect samples in CLIP latent space, allowing them to find the direction of failure modes.
However, \citet{jain2022distilling} aim to interpret failure modes as text prompts in the CLIP's embedding space, while \methodName{} focuses on generating black-box attacks on synthetic image detectors, by steering the latent space of a T2I generative model.

\textbf{Unrestricted Adversarial (UA) attacks} \citep{song2018constructing} on image classifiers directly use generative models to create attack images, in contrast to classical adversarial attacks that find a low-norm perturbation on clean images. Some early UA approaches \citep{song2018constructing,zhao2017generating,lapid2023patch,hu2021naturalistic} involved finding attack directions in the latent space of generative adversarial networks (GANs) \citep{goodfellow2014generative}. Recent generative models, such as diffusion models and flow matching models \citep{ho2020denoising,song2020score,lipman2022flow,liu2022flow} have also been employed for performing more sophisticated UA attacks \cite{chen2023content,xue2023diffusion,dai2024advdiff,chen2024diffusion,li2024advad,chen2023advdiffuser}.

\textbf{Diffusion-Based UA Attacks.}\quad Diffusion models are well known to be conditioned at inference time with gradients of arbitrary functions, an approach known as \emph{guidance} \cite{bansal2023universal}. To our knowledge, \citet{xue2023diffusion} first repurposed guidance for adversarial attacks by formulating a framework for projected gradient descent over the iterative denoising steps of diffusion models (DiffPGD). A concurrent work, AdvDiff \citep{dai2024advdiff}, performs a two-fold attack: by both guiding the diffusion trajectory with adversarial guidance, and optimizing the initial noise prior with the adversarial gradient until a successful attack is discovered. While DiffPGD and AdvDiff apply the adversarial gradient over several timesteps, an alternative approach, DiffAttack \citep{chen2024diffusion}, optimizes the diffusion latent only at one particular timestep, while also exploiting attention maps in the model architecture to preserve realness and structure.

\textbf{Research Gap.}\quad Despite their differences, we note that DiffPGD, AdvDiff, and DiffAttack all share a common feature: at inference time, they compute a image-specific adversarial gradient $\nabla_{\bm x_t} \mathcal{L}_{\text{attack}}(\hat{\bm x}_T; \bm{\theta})$, where $\bm{x}_t$ is a noisy sample at a timestep $t$, $\hat{\bm{x}}_T$ is a predicted clean image, and $\mathcal{L}_\text{attack}$ is an adversarial objective on some classifier $\bm{\theta}$. All these attacks would need complete access to the weights and gradients of an SID model, making them white-box attacks. In contrast, \methodName{} pre-computes a general attack direction from a \emph{distribution} of images, even with only hard label access. We also note that computing the gradient $\nabla_{\bm{x}_t} \mathcal{L}_{\text{attack}}(\hat{\bm{x}}_0; \bm{\theta})$ can be computationally intractable for large, billion-parameter T2I models and high-resolution images; \methodName{} bypasses this problem by (i) being a gradient-free, black-box method (\cref{sec:app})  and (ii) using resolution transferability (\cref{sec:app:res}).

\section{Concluding Remarks \label{sec:conclusion}}
In this paper, we present \methodName{}, a universal black-box \Rtd{} method for performing unrestricted adversarial (UA) attacks on synthetic image detectors (SID). To keep up with rapid advances in text-to-image (T2I) generative models,
\methodName{} performs the critical task of reinforcing SID models with new attacks and augmented data.
Due to its black-box nature,  \methodName{} is able to perform attacks on proprietary and commercial detectors that lead the field of deepfake and synthetic image detection.
Moreover, \methodName{} is a \emph{brew once, break many} approach: it computes a distribution-based (in contrast to image-specific) attack that is universally applicable over arbitrary images and different resolutions, enabling computationally efficient UA attacks.

\textbf{Limitations \& Future Directions.}\quad \methodName{} is based on a linear dependency between predicted realness and the T2I latents; in some cases, this dependency may be nonlinear.
We can use a nonlinear kernel in \cref{eq:U} to discover this dependency; however, this comes with the added complexity of learning a nonlinear decoder or solving a pre-image problem to map the nonlinear embedding space back to $\mathcal{Z}$. In addition, \methodName{} requires a limited hyperparameter search for appropriate values of $\lambda_t$, the details of which are expanded in the Appendix.

\textbf{Ethical Considerations.}\quad While adversarial generation is essential for evaluating and improving detector models' robustness, it can also be misused. \methodName{} is intended solely for responsible \Rt, and we strongly oppose its use for malicious purposes. Please refer to \cref{app:def} for defense mechanisms against malicious usage of \methodName{}.

\noindent\textbf{Acknowledgements:} Mashrur Morshed and Vishnu Boddeti are partially supported by the National Science Foundation (award \#2147116). Any opinions, findings, conclusions, or recommendations expressed in this material are those of the authors and do not necessarily reflect the views of the NSF.

{
\small
\bibliography{main}
\bibliographystyle{plainnat}
}

\newpage
\appendix

\section*{Appendix}

In our main paper, we propose \methodName{} for performing unrestricted adversarial attacks on synthetic image detectors.
Here, we provide some additional details to support our main results. The appendix section is structured as follows:

\begin{enumerate}
    \item Implementation Details in \cref{app:implement}
    \begin{enumerate}
        \item Threat Model in \cref{app:imp:tmodel}
        \item T2I Model Settings in \cref{app:imp:t2i}
        \item Computing Steering Directions in \cref{app:imp:steer}
        \item Hyperparameter Search for $\lambda_t$ in \cref{sec:app:hpsearch}
        \item Calibrating the Confidence Threshold of an SID in \cref{app:imp:cal}
        \item Implementation of Spectral Fingerprint Analysis in \cref{app:imp:spectral}
        \item Implementation of Projected Steering Directions in \cref{app:imp:proj}
        \item More Visualizations on the Effect of PolyJuice in Image Generation Process in \cref{app:imp:comp}
    \end{enumerate}
    \item Qualitative Analysis of Generated Attacks in \cref{app:qual}
    \begin{enumerate}
        \item Common Patterns in Successful Attacks in \cref{sec:app:common}
    \end{enumerate}
    \item Additional Results in \cref{app:add-results}
    \begin{enumerate}
        \item Spectral Fingerprint Analysis in \cref{app:res:spectral}
        \item Realness Shift in T2I Latent Space in \cref{app:res:shift}
        \item Additional image-space visualizations on the effect of \methodName{}\cref{app:res:image}
        \item Validating whether \methodName{} is applicable on diverse prompts in \cref{app:reb:diverse}
        \item Evaluating \methodName{} on additional SID models in \cref{app:reb:sid}
    \end{enumerate}
    \item Comparing \methodName{} against a Diffusion-based Transferred Attack in \cref{app:transfer}
    \item Potential Defense Mechanisms against \methodName{} in \cref{app:def}
\end{enumerate}

\section{Implementation Details \label{app:implement}}

\subsection{Threat Model \label{app:imp:tmodel}}

\begin{itemize}
  \item \textbf{Attacker’s Goal.}
  Given a text prompt $c$ and a latent $z$, the attacker aims to generate synthetic images using a text-to-image (T2I) generative model $G:\mathcal{Z}\times\mathcal{C}\rightarrow\mathcal{X}$ that deceive a target synthetic image detector (SID) $f:\mathcal{X}\rightarrow[0,1]$ into being misclassified as real (class $0$).

  \item \textbf{Attacker’s Capability.}
  \begin{itemize}
    \item \textbf{Black-box access to the SID:} The attacker can only query the SID and observe hard labels $Y$ (real/fake), without access to model weights or gradients.
    \item \textbf{Full access to the T2I generative model:} The attacker can manipulate the latent space $\mathcal{Z}$ and control the generation process of the T2I model.
    \item \textbf{Sufficient number of queries:} The attacker can generate a dataset of image–latent–label triplets to analyze the SID's behavior in response to various inputs.
  \end{itemize}

  \item \textbf{Attack Scenario.}
  \begin{enumerate}
    \item The attacker queries the black-box SID with fake images and obtains hard labels, constructing a dataset of TP and FN samples.
    \item The attacker pre-computes steering directions in the T2I latent space that correlate with an increased probability of being misclassified as real by the SID (\cref{eq:U}).
    \item At test time, the attacker applies this universal direction to arbitrary prompts, producing images that evade detection by the SID (\cref{eq:h}).
  \end{enumerate}
\end{itemize}

\subsection{T2I Model Settings \label{app:imp:t2i}}
We provide the generation settings for \sd{}, \dev{}, and \sch{} in \cref{tab:t2i-setting}. All models are available on Huggingface Models \citep{huggingface} with the corresponding Model ID provided in \cref{tab:t2i-setting}.

\begin{table}[h]
\centering
\caption{\footnotesize Settings for T2I Models.\label{tab:t2i-setting}}
\resizebox{\linewidth}{!}{%
\begin{tabular}{lcccccc}
        Model & Model ID & Num Steps & Guidance Scale & Max Seq. Length \\
        \midrule
        \sd{}      & \texttt{stabilityai/stable-diffusion-3.5-large} & 28  & 3.5  & 256  \\
        \dev{}     & \texttt{black-forest-labs/FLUX.1-dev} & 50  & 3.5  & 512 \\
        \sch{}     & \texttt{black-forest-labs/FLUX.1-schnell} & 4  & 0  &  256 \\
        \bottomrule
\end{tabular}%
}
\vspace{-0.3cm}
\end{table}

\subsection{Computing Steering Directions \label{app:imp:steer}}
In \cref{alg:steering}, we provide a pseudocode for calculating the steering direction $\bm \delta_t$ at a given timestep $t$. \cref{alg:steering} requires precomputed T2I latents $\bm Z_t \in \mathbb{R}^{N \times d_{\mathcal{Z}}}$, where $N$ is the number of generated images and $d_{\mathcal{Z}}$ is the dimensionality of the T2I latent space. For efficient calculation of the top-k eigenvalues and corresponding eigenvectors, we use the \texttt{LOBPCG} algorithm \citep{stathopoulos2002block}.

\begin{algorithm}
\caption{Compute Steering Direction at Timestep $t$\label{alg:steering}}
\begin{algorithmic}[1]
\Require $\bm Z_t \in \mathbb{R}^{N \times d_{\mathcal{Z}}}$, $\bm Y \in \mathbb{R}^{N \times 2}$ (one-hot SID predicted labels)
\State $\bm Z_t \gets \bm Z_t - \text{MEAN}(\bm Z_t, \text{dim}=0)$ \textcolor{gray!80!black}{\Comment{Center the data}}
\State $\bm C \gets\bm Z_t^\top \cdot \bm Y$ \textcolor{gray!80!black}{\Comment{Cross-covariance}}
\State $\bm A \gets \bm C \cdot \bm C^\top$ \textcolor{gray!80!black}{\Comment{Kernel matrix in \cref{eq:U}}}
\State $(\bm \sigma, \bm U) \gets \text{LOBPCG}(\bm A, k=2)$ \textcolor{gray!80!black}{\Comment{Top-2 eigenpairs}}
\State $\bm \delta_t \gets \bm \sigma_0 \cdot \bm U_0 + \bm \sigma_1 \cdot \bm U_1$ \textcolor{gray!80!black}{\Comment{Weighted steering direction in \cref{eq:delta}}}
\State \Return $\bm \delta_t$
\end{algorithmic}
\end{algorithm}

\subsection{Hyperparameter Search for $\lambda_t$ \label{sec:app:hpsearch}}

In \cref{eq:h}, \methodName{} steers a T2I latent $\bm Z_t$ using the steering direction $\bm \delta_t$ and a magnitude coefficient $\lambda_t \in [0, \infty)$.
In all steering approaches, finding the correct magnitude for adding the steering direction to the normal flow requires a hyperparameter search \cite{bansal2023universal,nair2023steered,friedrich2024auditing}, and \methodName{} is no exception. As a result, there is a need for finding an optimal $\lambda_t$.

\begin{figure}[h]
    \centering
    \includegraphics[width=\linewidth,height=0.4\linewidth]{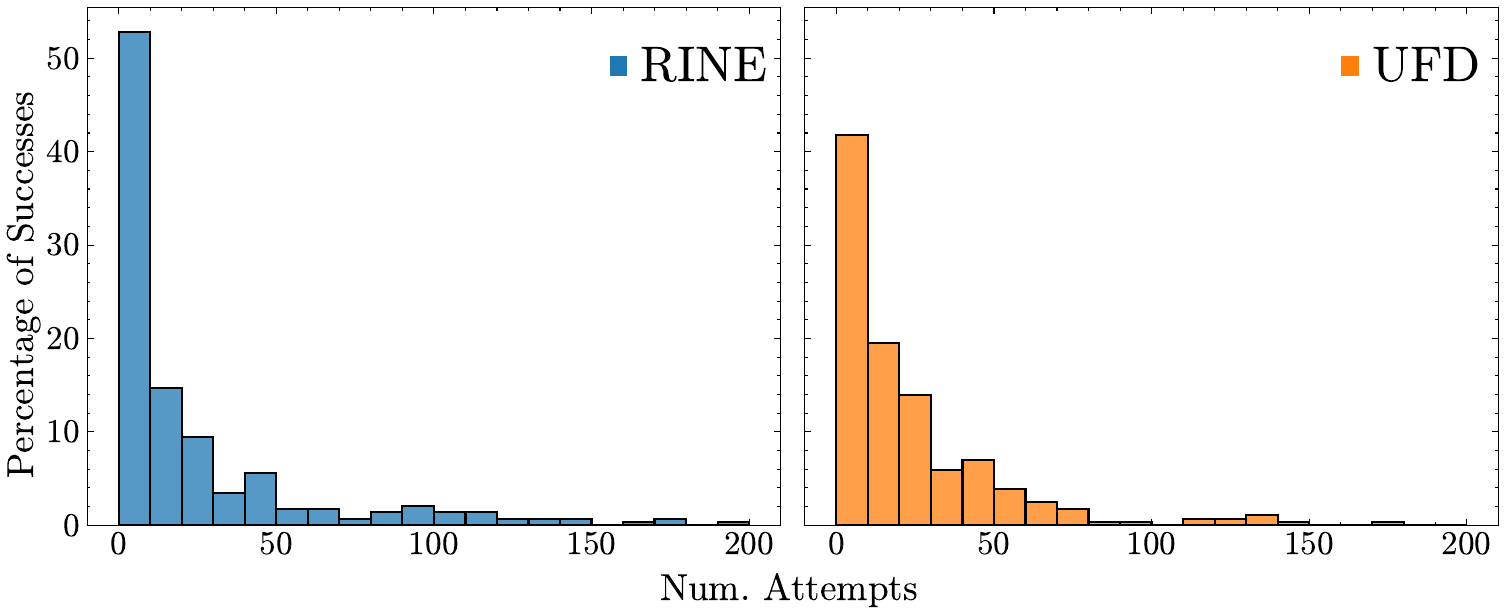}
    \caption{Number of attempts to find successful attacks using \dev{}. \label{fig:hist}}
    \vspace{-1em}
\end{figure}

To efficiently limit the search space for the set $\{\lambda_t\}_{t=0}^{T-1}$, we consider $\lambda_t = \lambda \cdot \mathbbm{1}\{ a \le t \le b\}$, that is, we only apply the steering directions $\bm \delta_t$ at a constant magnitude $\lambda$ over some continuous interval $[a, b] \subseteq [0, T)$. 
We then use a simple hyperparameter search based on the Optuna framework\citep{akiba2019optuna} to find $(\lambda, a, b)$. For any text caption $\bm c \in \mathcal{C}$, we define a \emph{budget}, or the maximum number of attempts that a T2I model can make to deceive an SID. We find that \methodName{} can find a majority of the successful attacks within few attempts (e.g. 10), as we show in \cref{fig:hist}. Further, for all captions, we generate all attacks with the same random seed (0) for determinism.

\begin{figure}[h]
    \centering
    \includegraphics[width=\linewidth,height=0.4\linewidth]{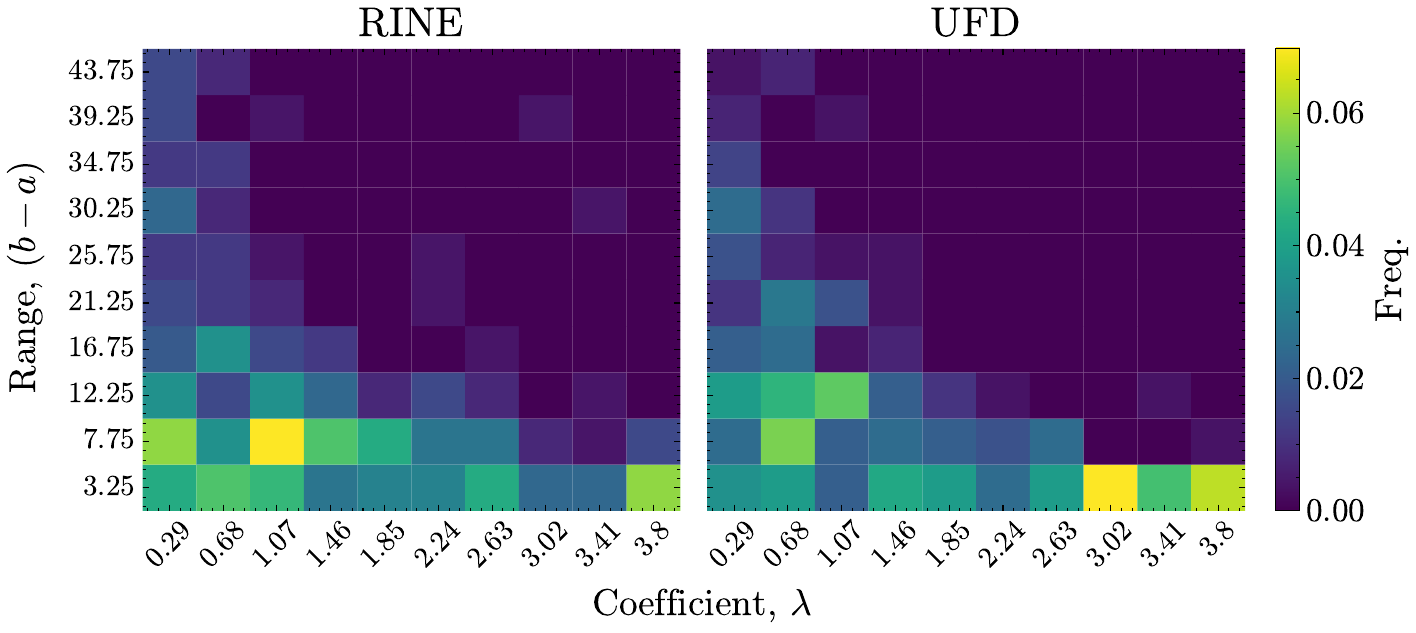}
    \caption{Heatmap showing $\lambda$ and corresponding range $(b-a)$ for \dev{}. \label{fig:heatmap}}
\end{figure}

\begin{figure}[h]
    \centering
    \includegraphics[width=\linewidth,height=0.4\linewidth]{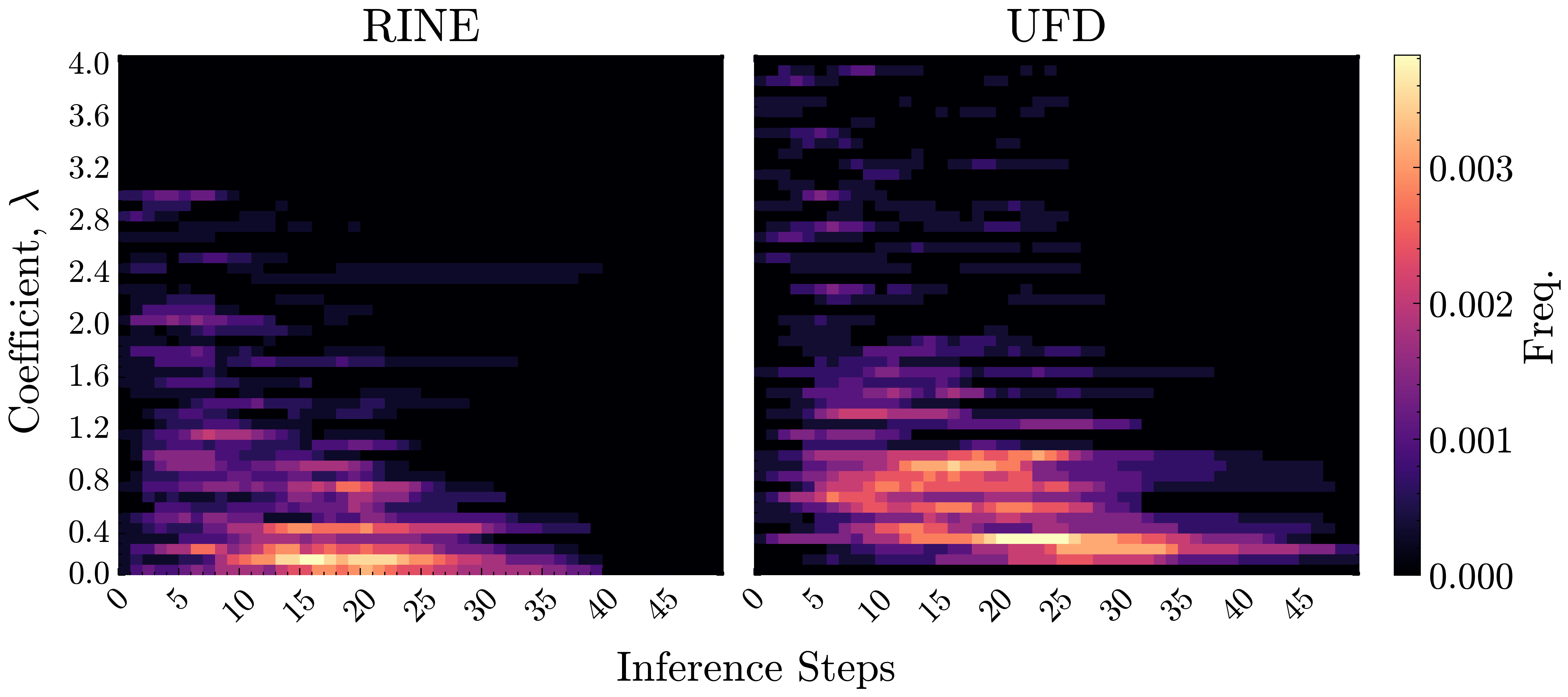}
    \caption{Heatmap showing $\lambda$ and \methodName{} steering steps for \dev{}.}
    \label{fig:heatmap_steps}
\end{figure}

We also analyze the hyperparameters corresponding to successful attacks as heatmaps.
From \cref{fig:heatmap}, we observe that there is a negative correlation between the constant magnitude coefficient $\lambda$ and the length of the continuous interval $[a, b]$ over which \methodName{} is applied. For higher values of $\lambda$, the steering direction $\bm \delta_t$ is applied to only a few steps, while for smaller values, the steering is done on a large number of steps.

Previously, \citet{esser2024scaling} show that intermediate timesteps are more important in the image generation process of flow matching-based T2I models.
This finding is supported by \cref{fig:heatmap_steps}, which depicts that \methodName{}-steering overall prefers intermediate steps and smaller coefficients.

\subsection{Calibrating the Confidence Threshold of an SID \label{app:imp:cal}}
For calibrating the threshold $\tau$ of SID models, we follow the algorithm used by \citet{ojha2023towards}. As shown in \cref{alg:thresh}, the algorithm requires the same number of \emph{real} and \emph{fake} samples and their corresponding predicted confidence scores as the input and return the value of the best threshold $\tau_{\text{best}}$ that maximizes the accuracy. In practice, for real images, we use a set of images from the training split of COCO, while the fake images are generated by T2I models. 

In \cref{tab:main-results}, we calibrate UFD using 20K real images and 20K generated images mixed from all the T2I models. 
For each T2I-SID pair in \cref{tab:ft-res}, we calibrate the SIDs (\includegraphics[width=0.8em]{figs/pj.png}-Cal. in \cref{tab:ft-res}) using the following setting. For $n$ successful \methodName{} attacks ($n < 1000$), we use $2n$ real images, $n$ synthetic images generated by the unsteered T2I models, and $n$ \methodName{}-steered images.

\begin{algorithm}[h]
\caption{Find the Best Threshold $\tau$ for SIDs (from \cite{ojha2023towards})
\label{alg:thresh}}
\begin{algorithmic}[1]
\Require Ground truth labels $y_\text{true}$, SID predicted scores $y_\text{pred}$ \textcolor{gray!80!black}{\Comment{where, $|y_\text{true} = 0| = |y_\text{true} = 1|$}}
\State $\text{indices} \gets \text{ARGSORT}(y_\text{true})$
\State $y_\text{true}, y_\text{pred} \gets y_\text{true}[\text{indices}], y_\text{pred}[\text{indices}]$ \textcolor{gray!80!black}{\Comment{Sort $y_\text{true}$ and $y_\text{pred}$ according to $y_\text{true}$}}
\State $N \gets \text{LEN}(y_\text{true})$
\If{$\max(y_{\text{pred}}[0:\lfloor N/2 \rfloor]) \leq \min(y_{\text{pred}}[\lfloor N/2 \rfloor:N])$} \textcolor{gray!80!black}{\Comment{Perfectly separable real and fake}}
    \State \Return $\frac{1}{2} \left( \max(y_{\text{pred}}[0:\lfloor N/2 \rfloor]) + \min(y_{\text{pred}}[\lfloor N/2 \rfloor:N]) \right)$
\EndIf
\State $\text{best\_acc}, \tau_\text{best} \gets 0, 0$
\ForAll{$\tau \in y_{\text{pred}}$} \textcolor{gray!80!black}{\Comment{Greedily test each $y_\text{pred}$ as a threshold}}
    \State $\text{temp} \gets y_{\text{pred}}$
    \For{$i = 0$ to $N-1$}
    \State $\text{temp}[i] \gets \mathbbm{1}\bigl\{\text{temp}[i] \geq \tau\bigr\}$
    \EndFor
    \State $\text{acc} \gets \frac{1}{N} \sum_{i=0}^{N-1} \mathbbm{1}\bigl\{\text{temp}[i] = y_{\text{true}}[i]\bigr\}$
    \If{$\text{acc} \geq \text{best\_acc}$}
        \State $\tau_\text{best} \gets \tau$
        \State $\text{best\_acc} \gets \text{acc}$
    \EndIf
\EndFor
\State \Return $\tau_\text{best}$
\end{algorithmic}
\end{algorithm}

\subsection{Implementation of Spectral Fingerprint Analysis \label{app:imp:spectral}}

Given a generated image $\bm x^\text{(i)}$, we assume that the high-frequency details in $\bm x^\text{(i)}$ are a sum of (i) a deterministic component arising from the generative model, i.e. the fingerprint $F$, and (ii) a random component $\bm w^\text{(i)} \sim \mathcal{N}(\bm 0, \bm I)$ \citep{corvi2023detection}.

For each group of images, we first apply a denoising filter and estimate a noise residual by computing the difference from the original image. Then we visualize the average residuals in the frequency domain, as shown in \Cref{fig:freq,fig:app:freq}. The algorithm for computing spectral fingerprints is provided in \cref{alg:freq}.

\begin{algorithm}[h]
\caption{Computing Spectral Fingerprint of Images \label{alg:freq}}
\begin{algorithmic}[1]
\Require A set of images $X$
\State $X_{\text{denoised}} \gets \text{NL-Mean}(X)$ \textcolor{gray!80!black}{\Comment{Denoise the images}}
\State $X_{\text{Residual}} \gets X - X_{\text{denoised}} $ \textcolor{gray!80!black}{\Comment{Compute the residuals}}
\State $F \gets \text{FFT}(X_{\text{Residual}})$  \textcolor{gray!80!black}{\Comment{Perform Fast Fourier Transform}}
\State $S \gets \| \text{FFT\_SHIFT}(F) \|$ \textcolor{gray!80!black}{\Comment{Shift the zero-frequency component to the center of the spectrum}}
\State $S_{\log} \gets \log(1 + S)$ 
\State \Return $S_{\log}$
\end{algorithmic}
\end{algorithm}

\subsection{Implementation of Projected Steering Directions \label{app:imp:proj}}

\cref{fig:vector} shows the update directions of \methodName{}-steered T2I models in a projected subspace. For a given timestep $t$, we first solve the eigenvalue problem associated with \cref{eq:U}, to obtain an orthogonal projection matrix $\bm U_t$. The \methodName{}-steering direction at the current timestep, $\bm \delta_t$ is computed from a convex combination over the columns of $\bm U_t$, as described by \cref{eq:delta}.

Next, for each latent $\bm z_t$, we map the latent to the subspace by computing $\bm U^\top_t \bm z_t$. We also consider the latent at the next timestep $t + 1$ and similarly project it as $\bm U_{t}^\top \bm z_{t+1}$. Subsequently, the update vector to the next step can be defined as $\bm u_t = \bm U_t^\top (\bm z_{t + 1} - \bm z_t)$.

For both unsteered and \methodName{}-steered T2I latents, we plot the unit update direction $\bm \hat{u}_t$, positioned at the corresponding mapped points $\bm U_t^\top \bm z_t$.

\subsection{Experiments Compute Resources \label{app:imp:comp}}
For all experimental steps—including dataset generation, direction computation, and attacks—we used eight NVIDIA RTX A6000 GPUs, each with 48 GB of memory. The primary computational bottleneck arises from the memory requirements of the T2I models during image generation; \methodName{} itself adds negligible overhead. For the highest image resolution considered in this paper, image generation consumed approximately 75\% of GPU memory, equivalent to 36 GB.

\newpage

\section{Qualitative Analysis of Generated Attacks \label{app:qual}}

\begin{figure}[h!]
    \centering
    \includegraphics[width=\linewidth]{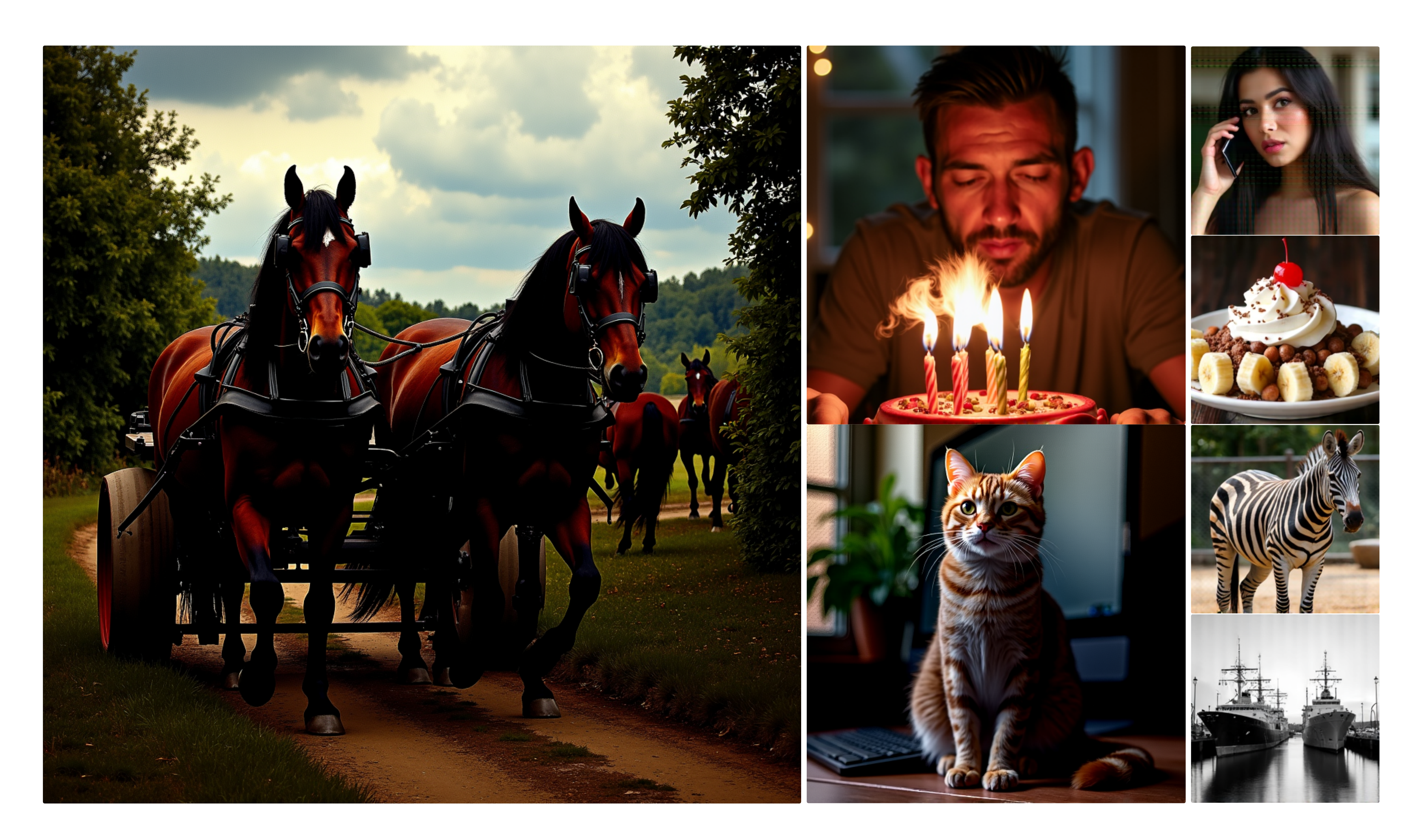}
    \caption{Attacks generated by \methodName{}-steered \dev{} model that were able to deceive RINE.}
    \label{fig:dev-rine-qua}
\end{figure}

\begin{figure}[h!]
    \centering
    \includegraphics[width=\linewidth]{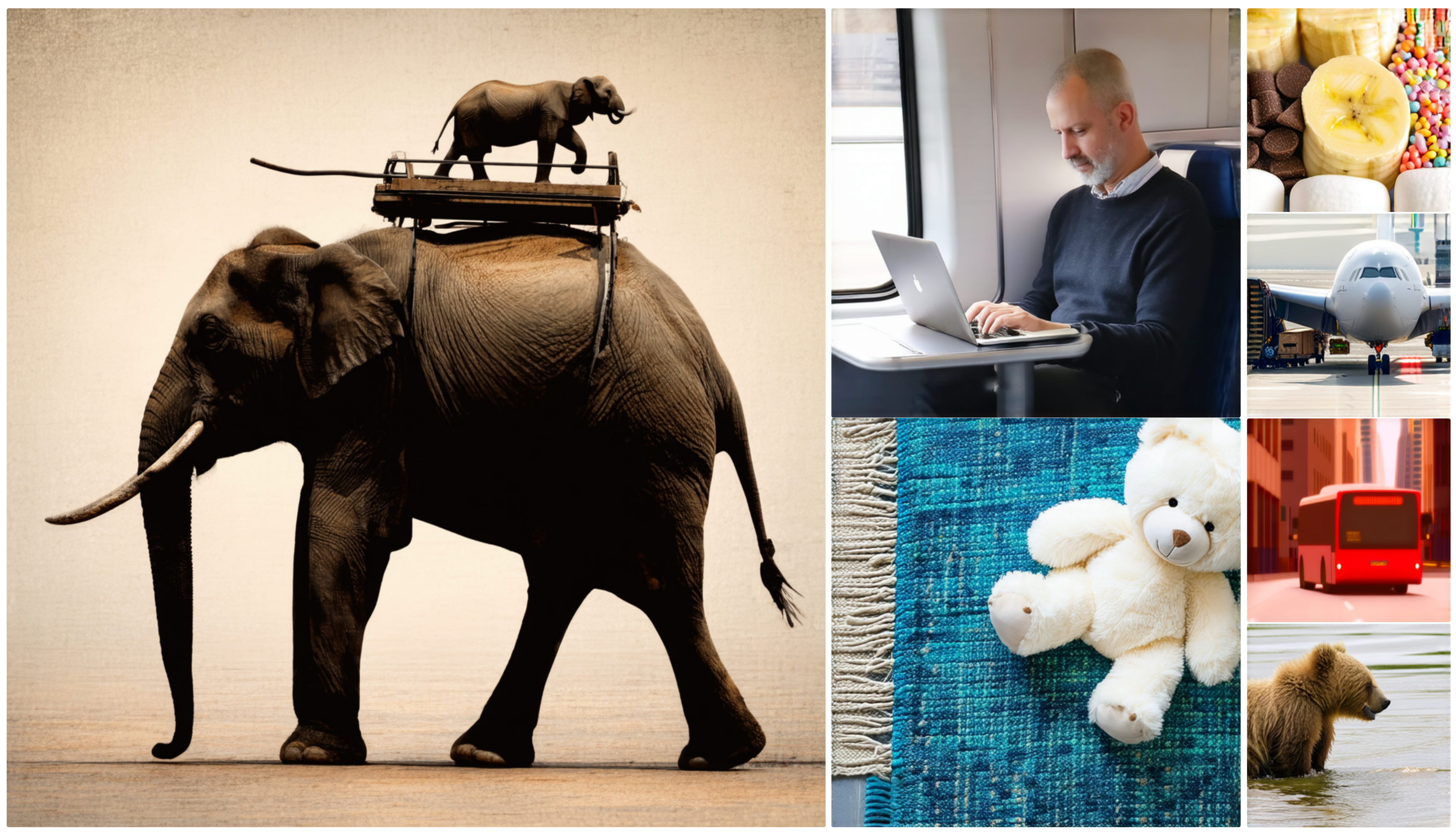}
    \caption{Attacks generated by \methodName{}-steered \sd{} model that were able to deceive UFD.}
    \label{fig:sd-ufd-qua}
\end{figure}

\begin{figure}[h!]
    \centering
    \includegraphics[width=\linewidth]{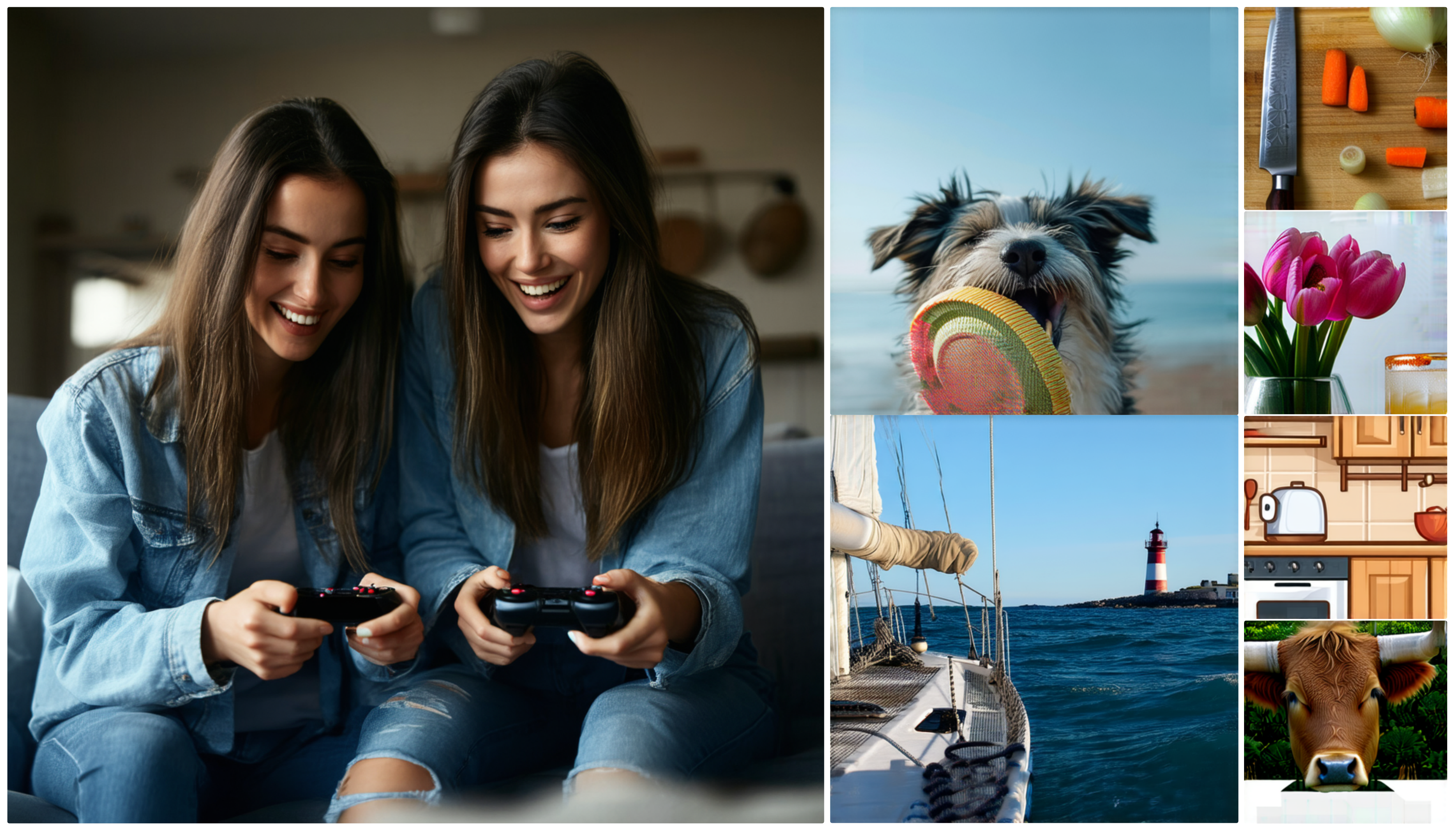}
    \caption{Attacks generated by \methodName{}-steered \sd{} model that were able to deceive RINE.}
    \label{fig:sd-rine-qua}
\end{figure}

\begin{figure}[h!]
    \centering
    \includegraphics[width=\linewidth]{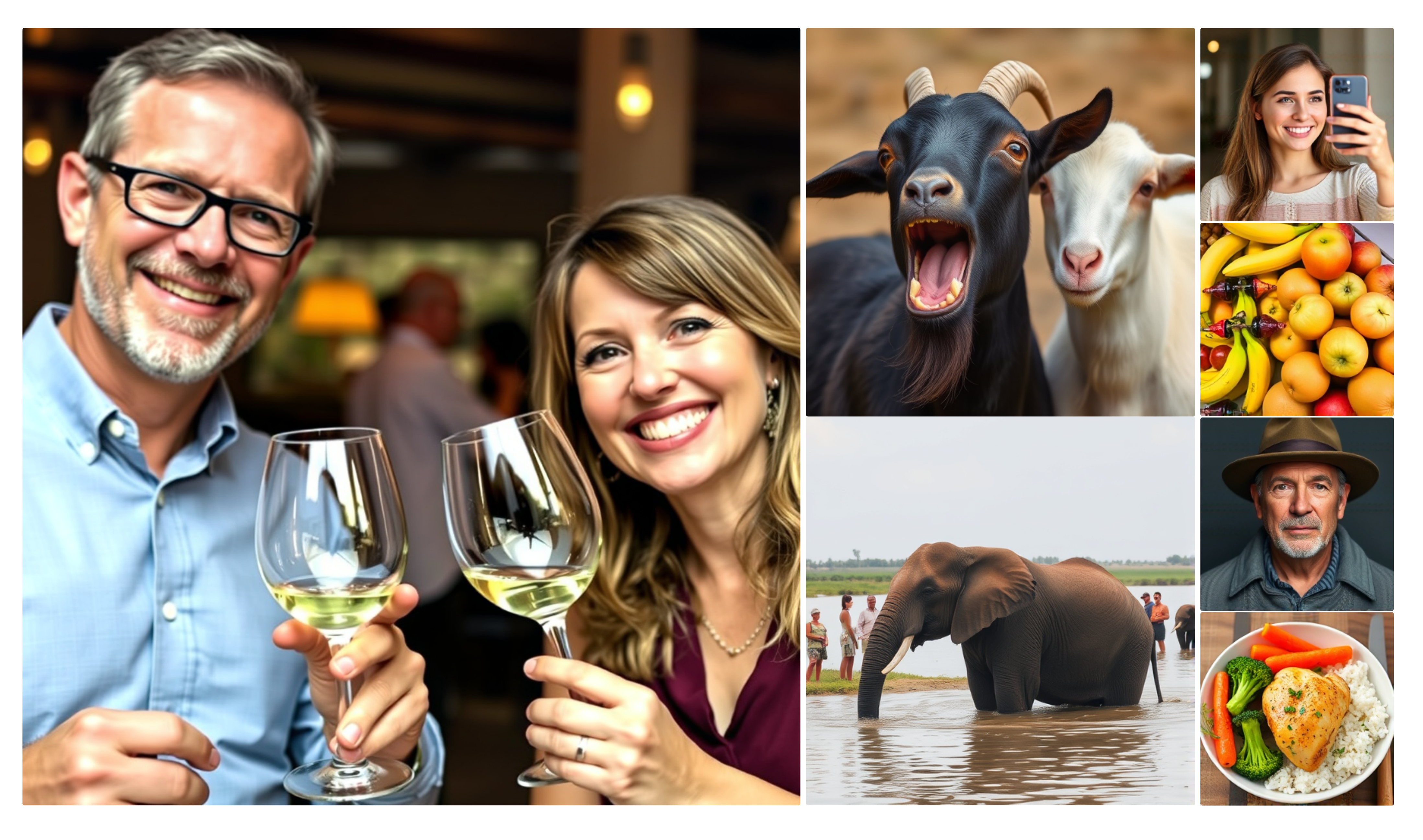}
    \caption{Attacks generated by \methodName{}-steered \sch{} model that were able to deceive UFD.}
    \label{fig:sch-ufd-qua}
\end{figure}

\begin{figure}[h!]
    \centering
    \includegraphics[width=\linewidth]{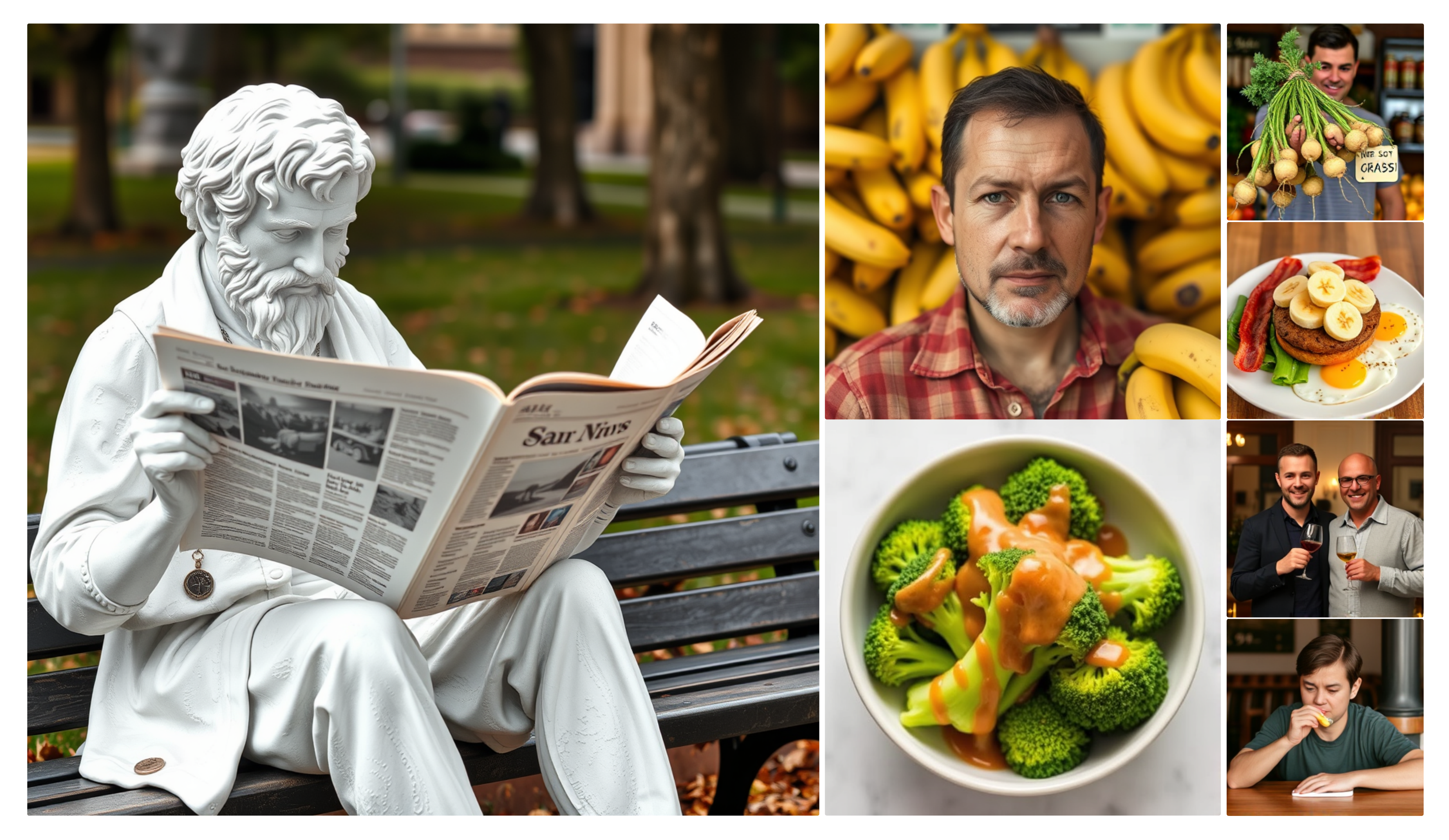}
    \caption{Attacks generated by \methodName{}-steered \sch{} model that were able to deceive RINE.}
    \label{fig:sch-rine-qua}
\end{figure}

\begin{figure}[h!]
    \centering
    \includegraphics[width=\linewidth]{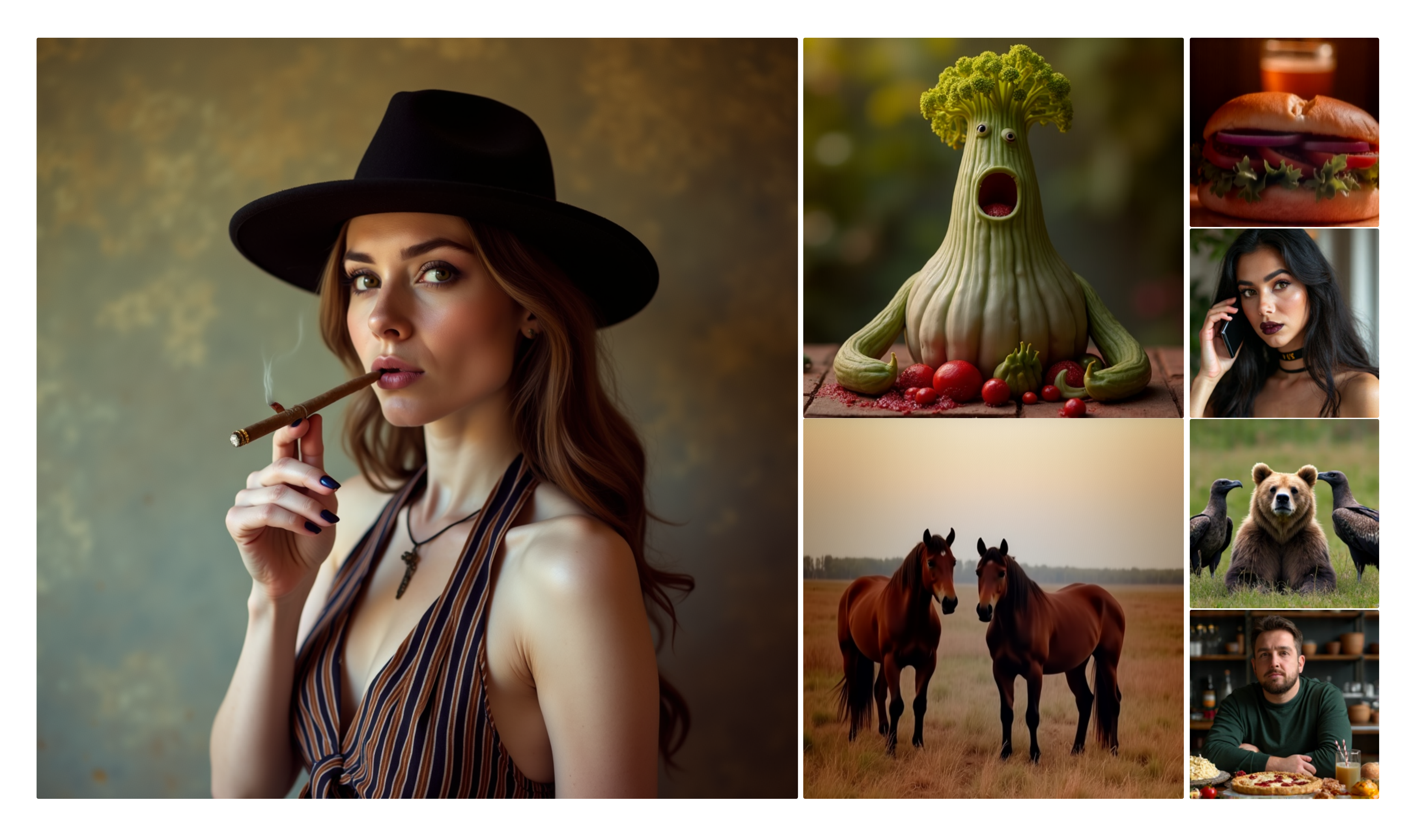}
    \caption{Attacks generated by \methodName{}-steered \dev{} model that were able to deceive UFD.}
    \label{fig:dev-ufd-qua}
\end{figure}

We provide some additional qualitative examples of successful attacks from \methodName{}-steered T2I models in \cref{fig:dev-rine-qua,fig:dev-ufd-qua,fig:sch-rine-qua,fig:sch-ufd-qua,fig:sd-rine-qua,fig:sd-ufd-qua}. In general, most of the images look realistic, even though we do not explicitly enforce any realism constraint. However, we notice that there are some characteristics of \methodName{}-generated attacks against specific SID models, which we discuss later in \cref{sec:app:common}.

\clearpage

\subsection{Common Patterns in Successful Attacks \label{sec:app:common}}
\begin{figure}[h]
    \centering
    \includegraphics[width=\linewidth]{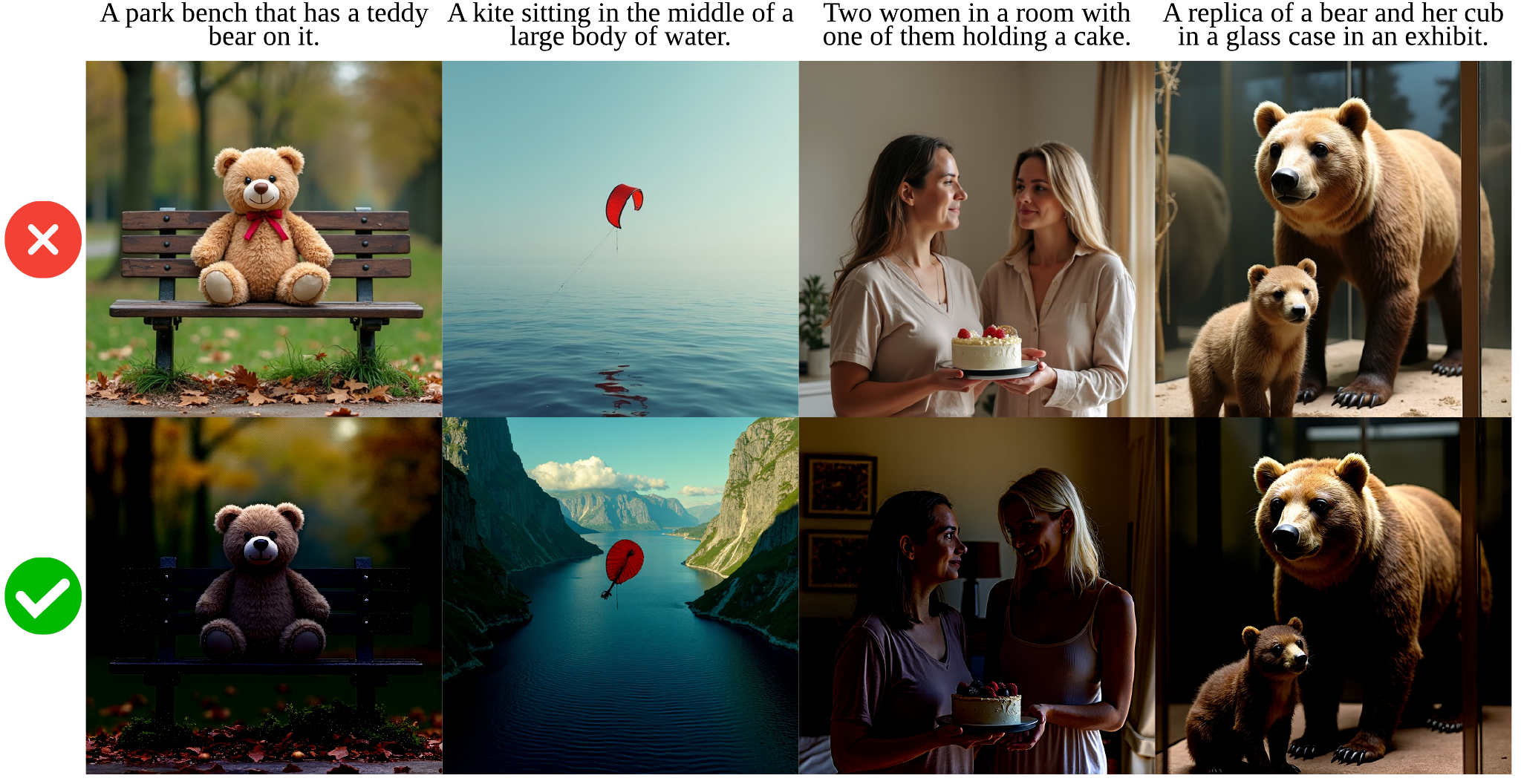}
    \caption{(Top) Unsteered T2I-generated images that RINE correctly detects as fake. (Bottom) \methodName{}-steered images that successfully deceive RINE as real. \label{fig:rine-dark}}
\end{figure}
\textbf{Common Patterns in Successful Attacks on RINE.}\quad In successful attacks against RINE, we observe that the generated images often exhibit low brightness.
In \cref{fig:rine-dark}, we show some images generated by unsteered (top) and \methodName{}-steered (bottom) T2I models, using captions from the COCO validation set. Further, the unsteered and steered images share the same random initial latent (i.e. they are all generated with the random seed 0).
Although RINE successfully detects the unsteered images as fake, it is deceived by the relatively \emph{darker} \methodName{}-steered images. This suggests a vulnerability of RINE to synthetic images that appear underexposed or have low brightness levels.

\begin{figure}[h]
    \centering
    \includegraphics[width=\linewidth]{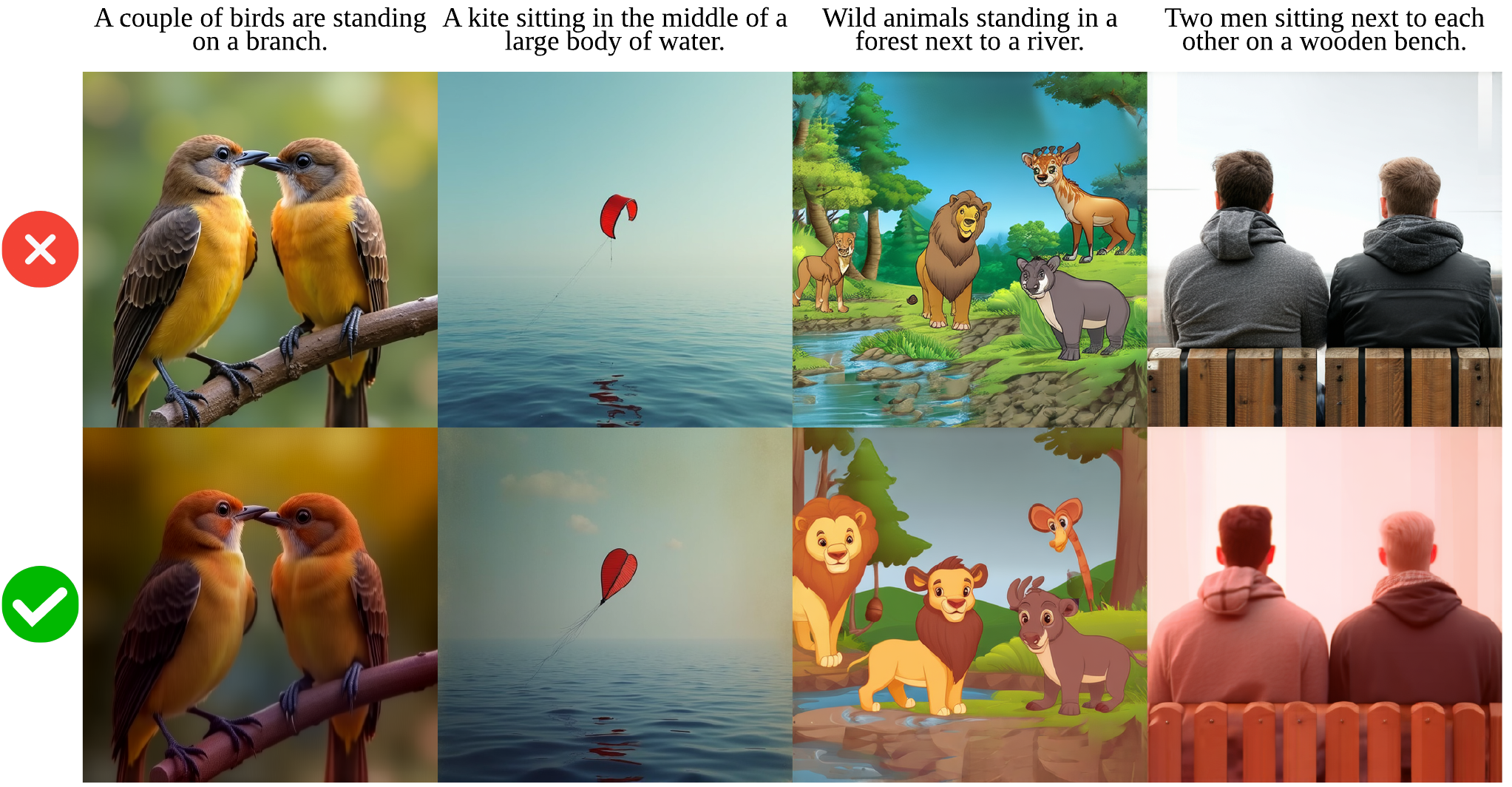}
    \caption{(Top) Unsteered T2I-generated images that UFD correctly detects as fake. (Bottom) \methodName{}-steered images that successfully deceive UFD as real. \label{fig:ufd-warm}}
\end{figure}

\textbf{Common Patterns in Successful Attacks on UFD.}\quad  In successful attacks against UFD, we observe that the generated images often exhibit warm colors. \cref{fig:ufd-warm} shows some examples unsteered images (top) that UFD detects as fake, and the corresponding \methodName{}-steered images (bottom) that fools the detector. Even on obviously fake images, such as the third column in \cref{fig:ufd-warm} (cartoon of wild animals, generated by \sd{}), \methodName{} produces an image with a warmer tone that UFD cannot detect as fake. This suggests a vulnerability of UFD to synthetic images with warm color temperatures.

\section{Additional Results \label{app:add-results}}

\subsection{Spectral Fingerprint Analysis \label{app:res:spectral}}

\begin{figure}[h]
    \centering
    \begin{subfigure}[t]{0.24\linewidth}
        \includegraphics[width=\linewidth]{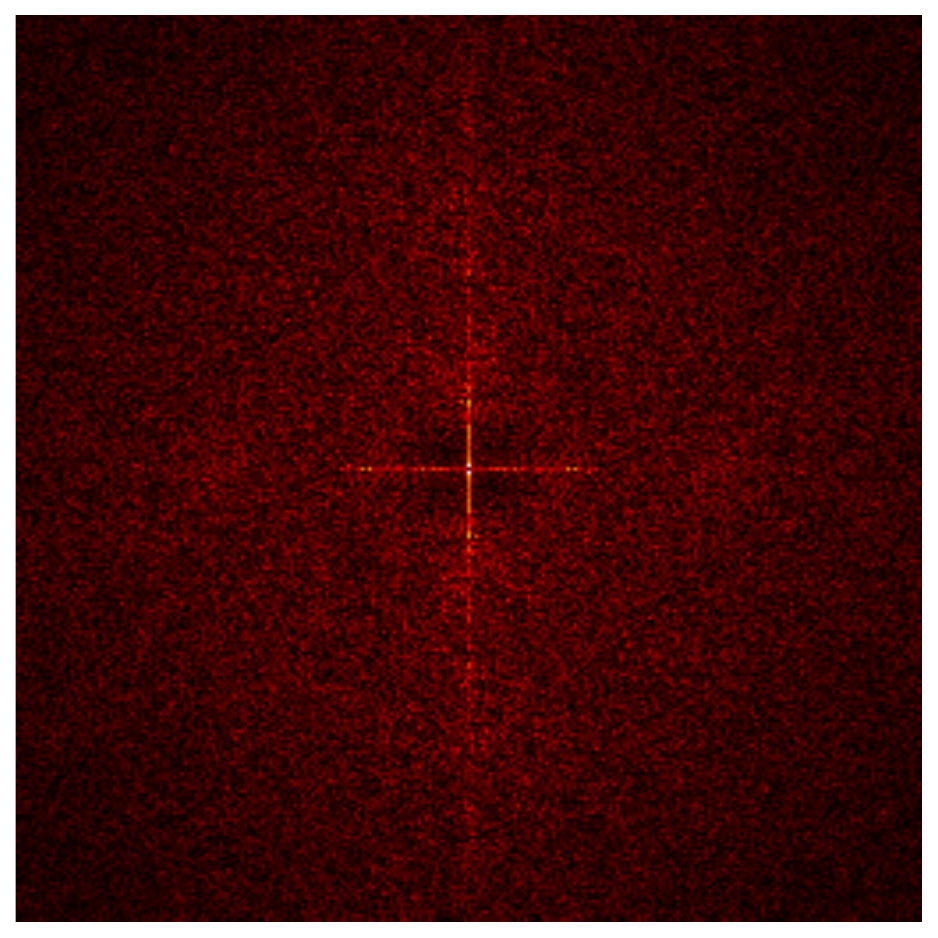}
        \caption{Real Samples\label{fig:app:freq:real:dev}}
    \end{subfigure}
    \hfill
    \begin{subfigure}[t]{0.24\linewidth}
        \includegraphics[width=\linewidth]{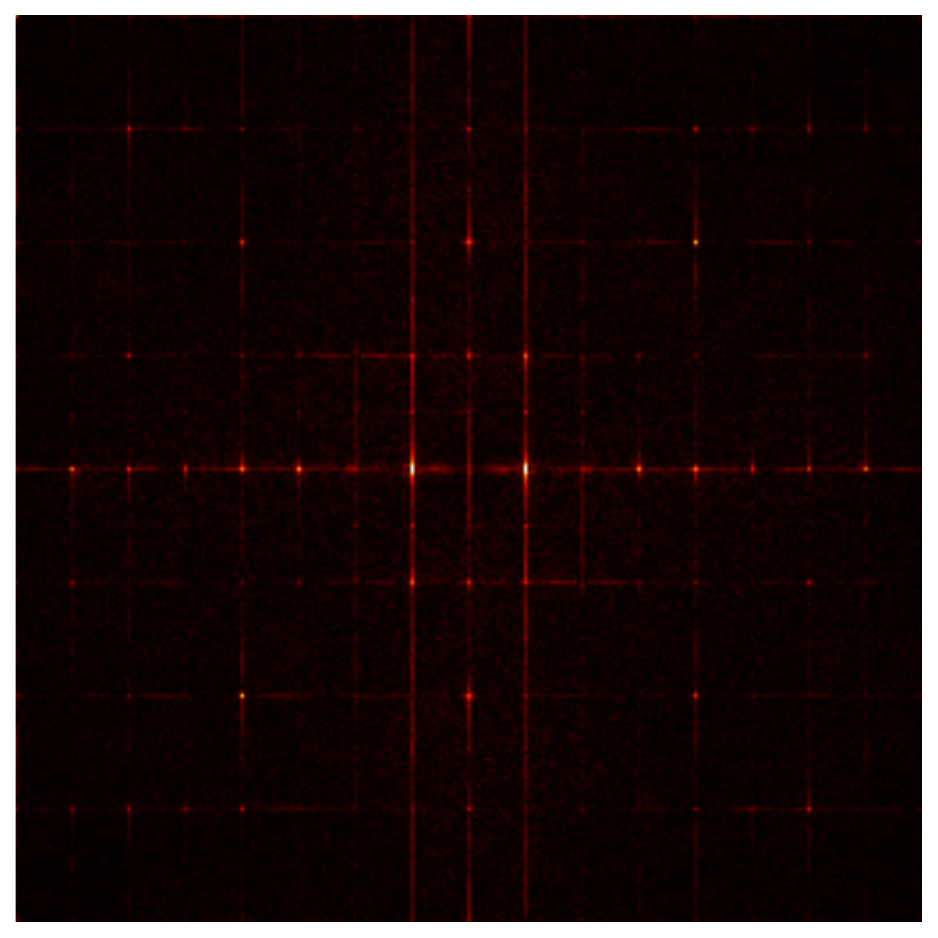}
        \caption{TP Samples\label{fig:app:freq:tp:dev}}
    \end{subfigure}
    \hfill
    \begin{subfigure}[t]{0.24\linewidth}
        \includegraphics[width=\linewidth]{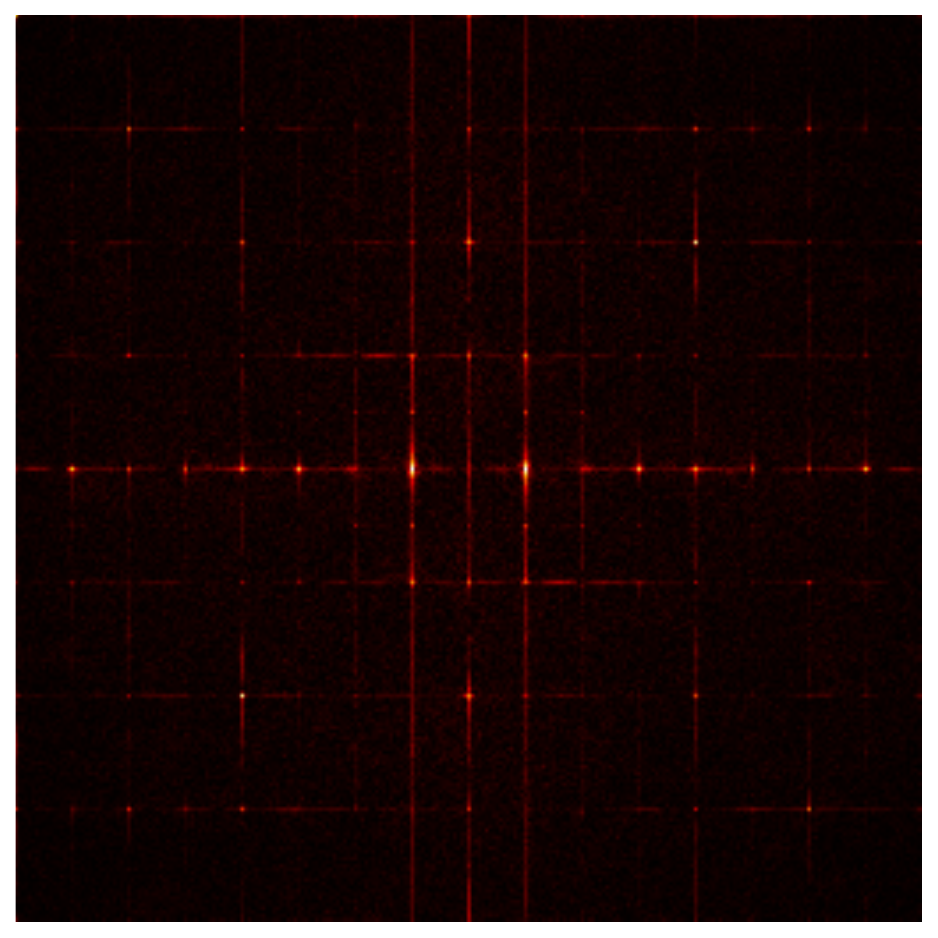}
        \caption{FN Samples\label{fig:app:freq:fn:dev}}
    \end{subfigure}
    \hfill
    \begin{subfigure}[t]{0.24\linewidth}
        \includegraphics[width=\linewidth]{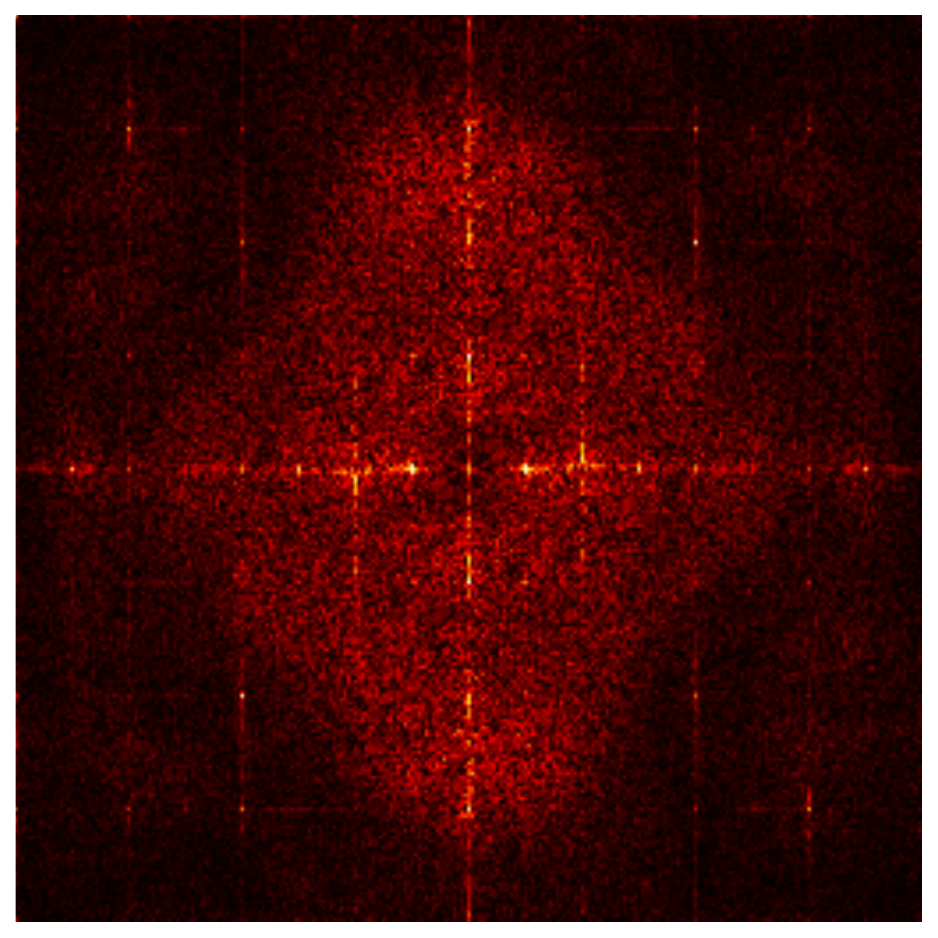}
        \caption{\methodName{} Attacks\label{fig:app:freq:att:dev}}
    \end{subfigure}
    \\
    \begin{subfigure}[t]{0.24\linewidth}
        \includegraphics[width=\linewidth]{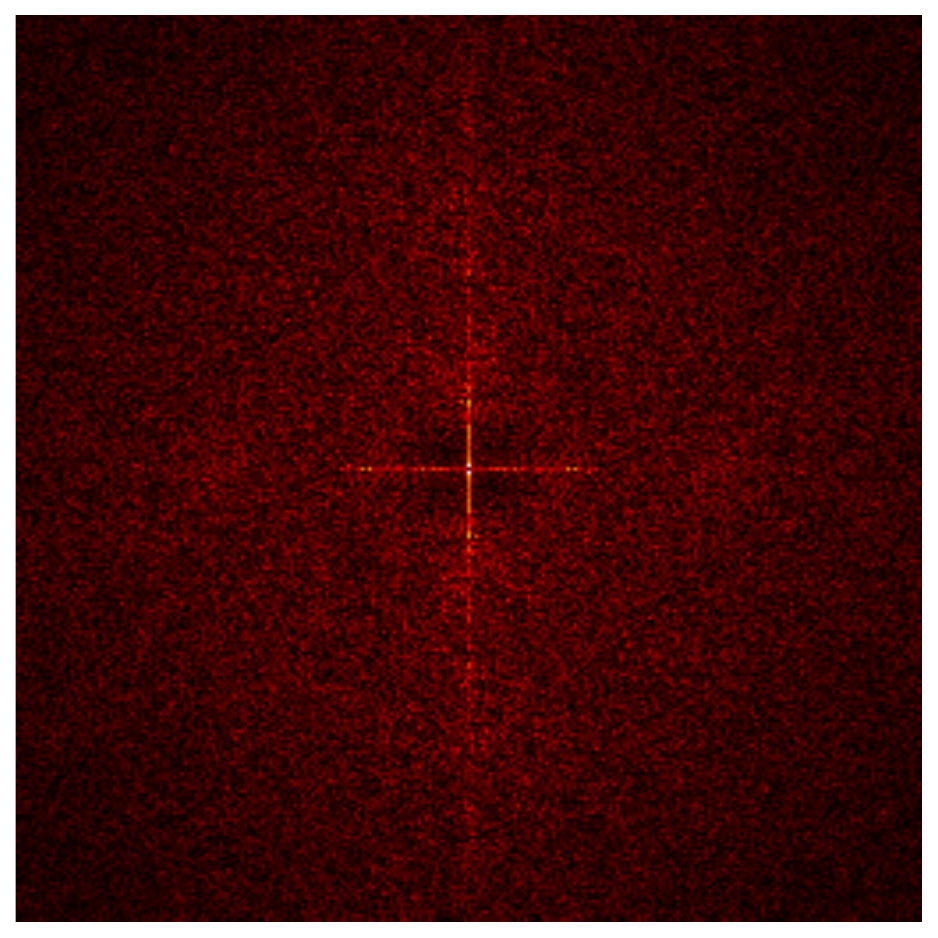}
        \caption{Real Samples\label{fig:app:freq:real:sch}}
    \end{subfigure}
    \hfill
    \begin{subfigure}[t]{0.24\linewidth}
        \includegraphics[width=\linewidth]{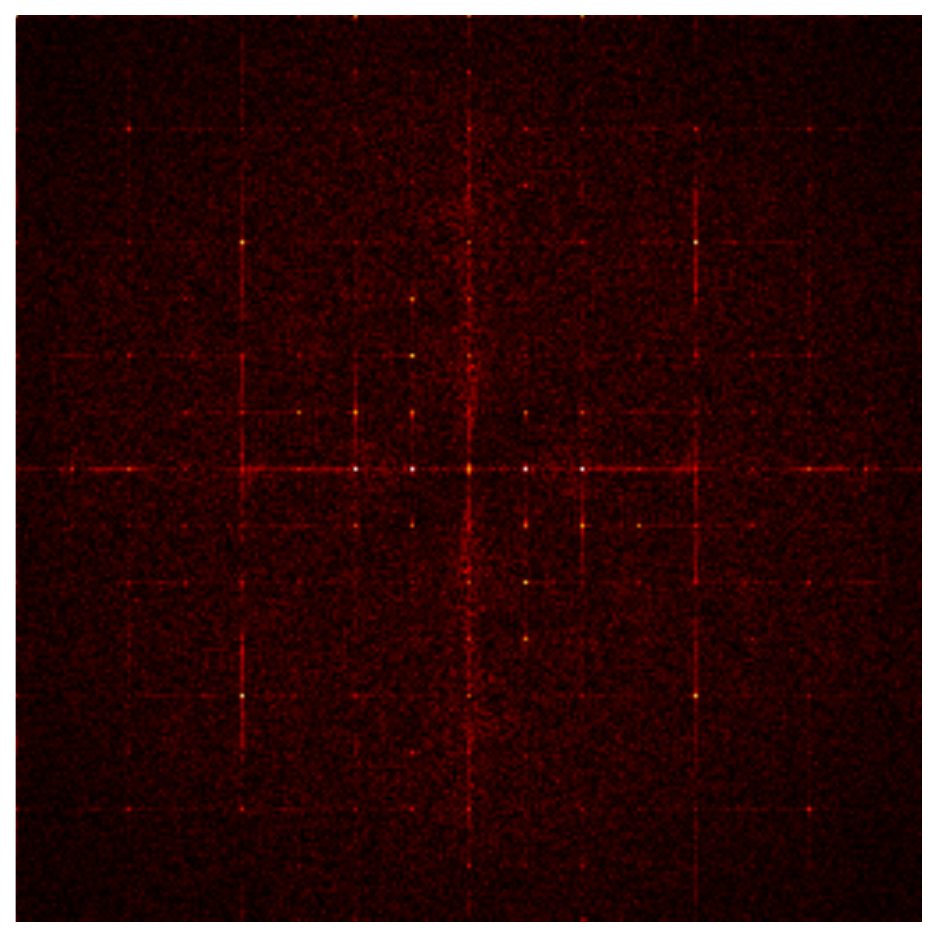}
        \caption{TP Samples\label{fig:app:freq:tp:sch}}
    \end{subfigure}
    \hfill
    \begin{subfigure}[t]{0.24\linewidth}
        \includegraphics[width=\linewidth]{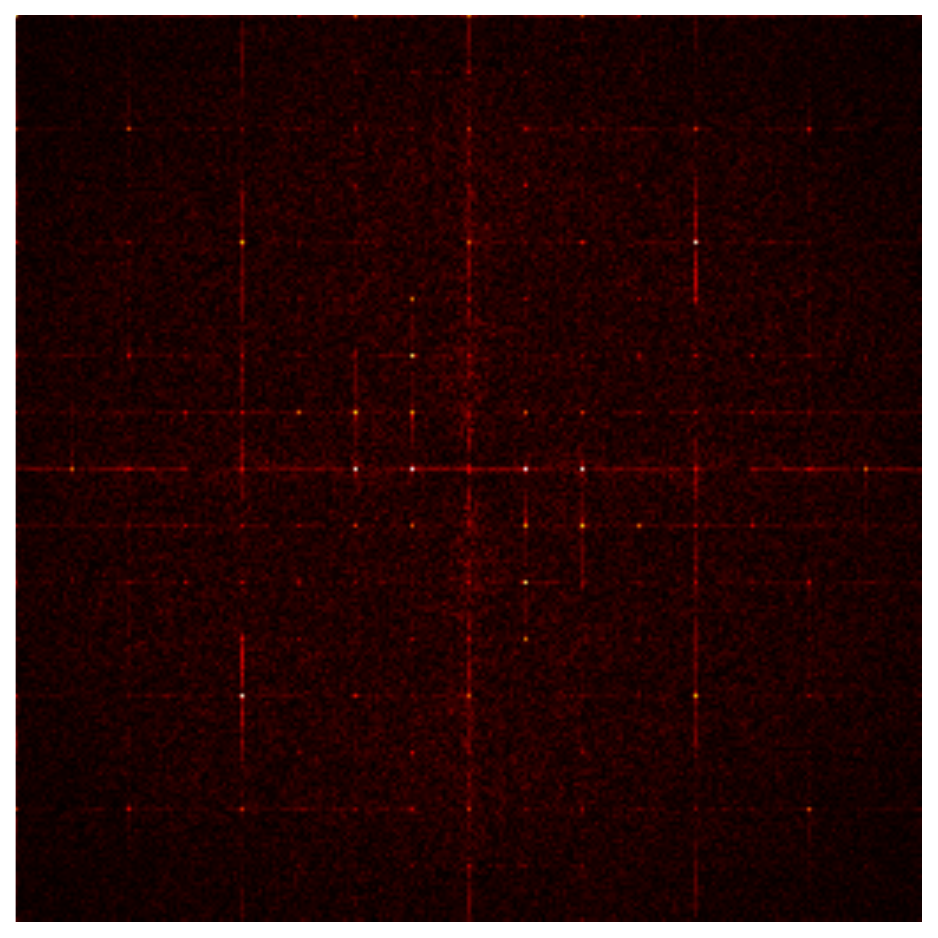}
        \caption{FN Samples\label{fig:app:freq:fn:sch}}
    \end{subfigure}
    \hfill
    \begin{subfigure}[t]{0.24\linewidth}
        \includegraphics[width=\linewidth]{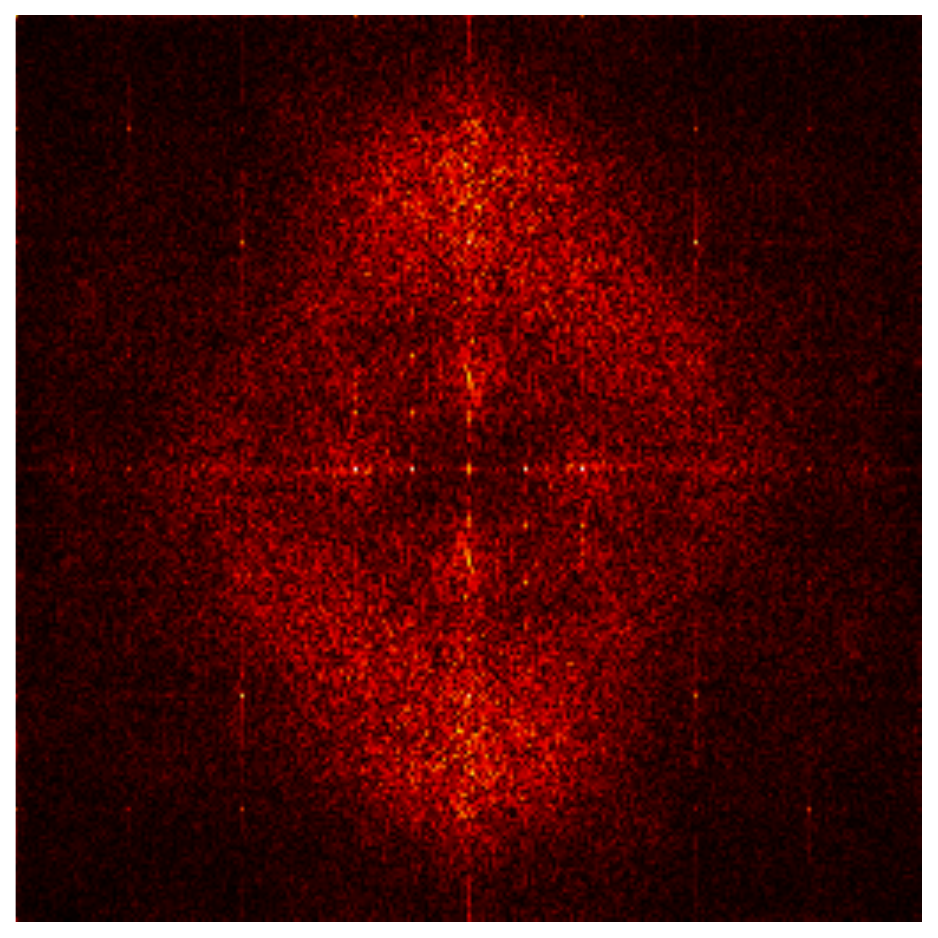}
        \caption{\methodName{} Attacks\label{fig:app:freq:att:sch}}
    \end{subfigure}
    \caption{More results on \textbf{Average Frequency Spectra} of COCO images and generated counterparts where $1^{st}$ and $2^{nd}$ rows correspond to samples from \dev{} and \sch{}, respectively.}
    \label{fig:app:freq}
\end{figure}

\cref{fig:freq} in the main paper illustrates the spectral fingerprints of real, unsteered, and \methodName{}-steered samples generated by \sd{}. Here, we extend the analysis to the other two T2I models, \dev{} and \sch{}. The first and second rows of \cref{fig:app:freq} show the spectral fingerprints of samples generated by \dev{} and \sch{}, respectively. We observe that \methodName{}-steered attacks effectively obscure the characteristic frequency patterns of their underlying T2I models, producing spectra that more closely resemble those of real images compared to unsteered attacks.

\subsection{Realness Shift in T2I Latent Space \label{app:res:shift}}
\begin{figure}[h]
    \centering
    \begin{subfigure}[t]{0.24\linewidth}
        \includegraphics[width=\linewidth]{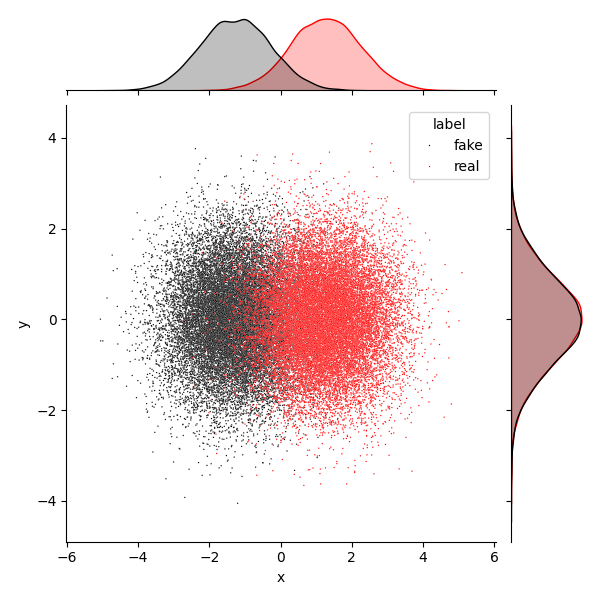}
        \caption{\hspace{3em}}
    \end{subfigure}
    \begin{subfigure}[t]{0.24\linewidth}
        \includegraphics[width=\linewidth]{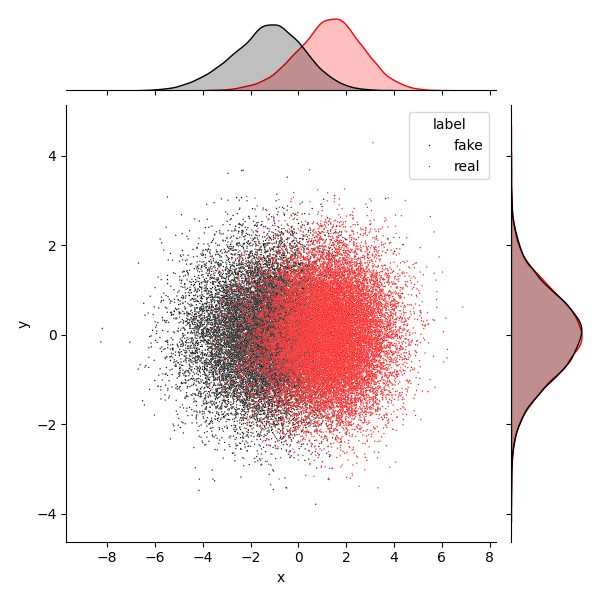}
        \caption{\hspace{3em}}
    \end{subfigure}
    \begin{subfigure}[t]{0.24\linewidth}
        \includegraphics[width=\linewidth]{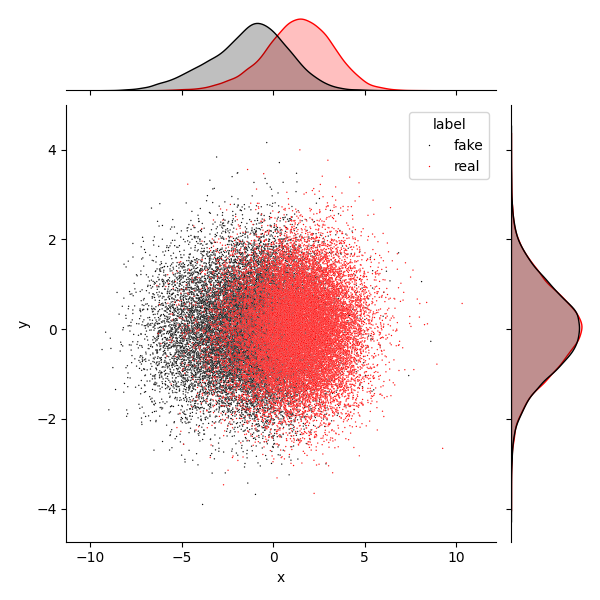}
        \caption{\hspace{3em}}
    \end{subfigure}
    \begin{subfigure}[t]{0.24\linewidth}
        \includegraphics[width=\linewidth]{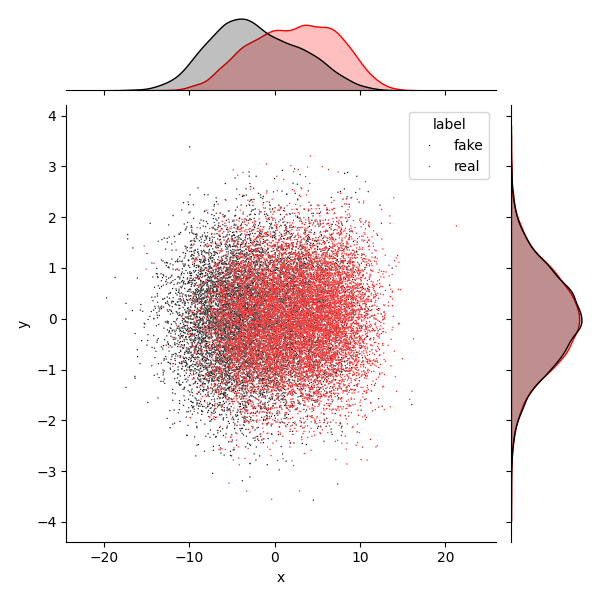}
        \caption{\hspace{3em}}
    \end{subfigure}
    \caption{More visualizations of the distribution shift in the T2I model's latent space.}
    \label{fig:app:shift}
\end{figure}
In \cref{fig:teaser:shift} of the main paper, we showed the realness shift in the latent space of T2I model for one timestep. In \cref{fig:app:shift}, we show the shift for four timesteps. We observe that there exists a clear shift between the predicted real and fake samples. However, the degree of linear separability of these distributions is not constant across different timesteps.

\subsection{More Visualizations on the Effect of \methodName{} in Image Generation Process \label{app:res:image}}
\begin{figure}[h]
    \centering
    \begin{subfigure}[t]{\linewidth}
        \centering
        \includegraphics[width=\linewidth]{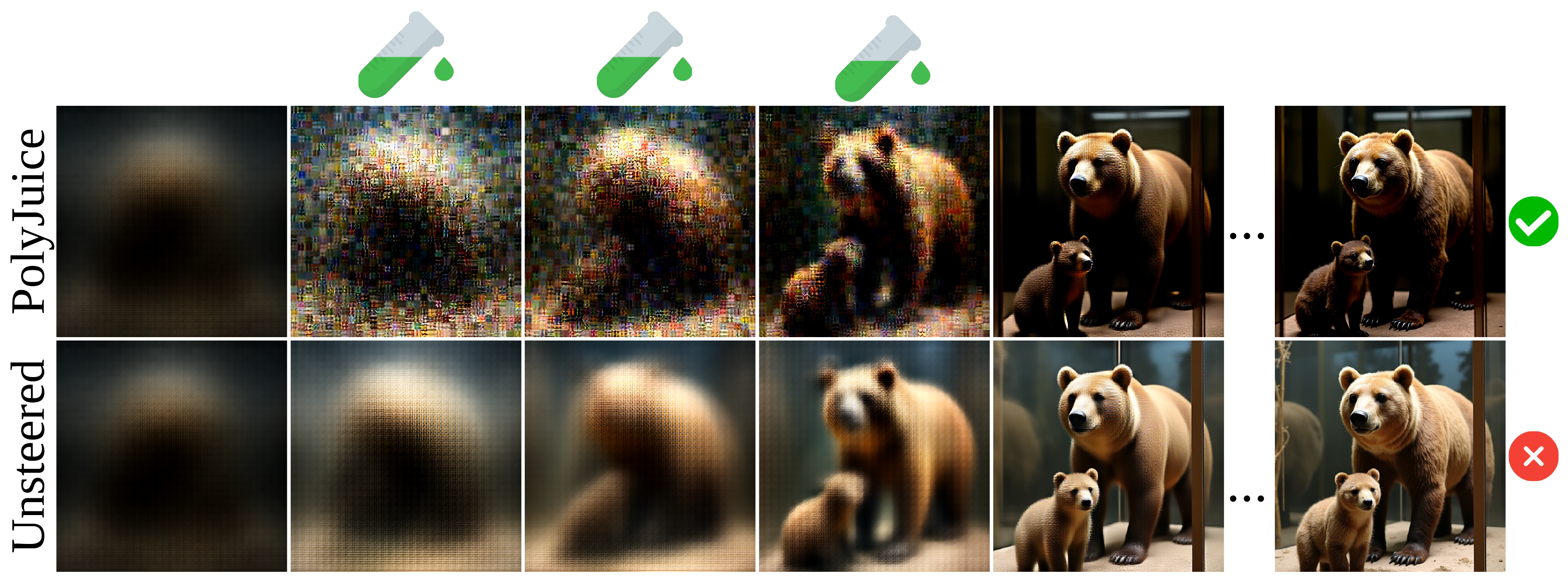}
        \caption{Intermediate steps for ``A replica of a bear and her cub in a glass case in an exhibit.''}
        \label{fig:app:steer-a}
    \end{subfigure}
    
    \vspace{1em} %

    \begin{subfigure}[b]{\linewidth}
        \centering
        \includegraphics[width=\linewidth]{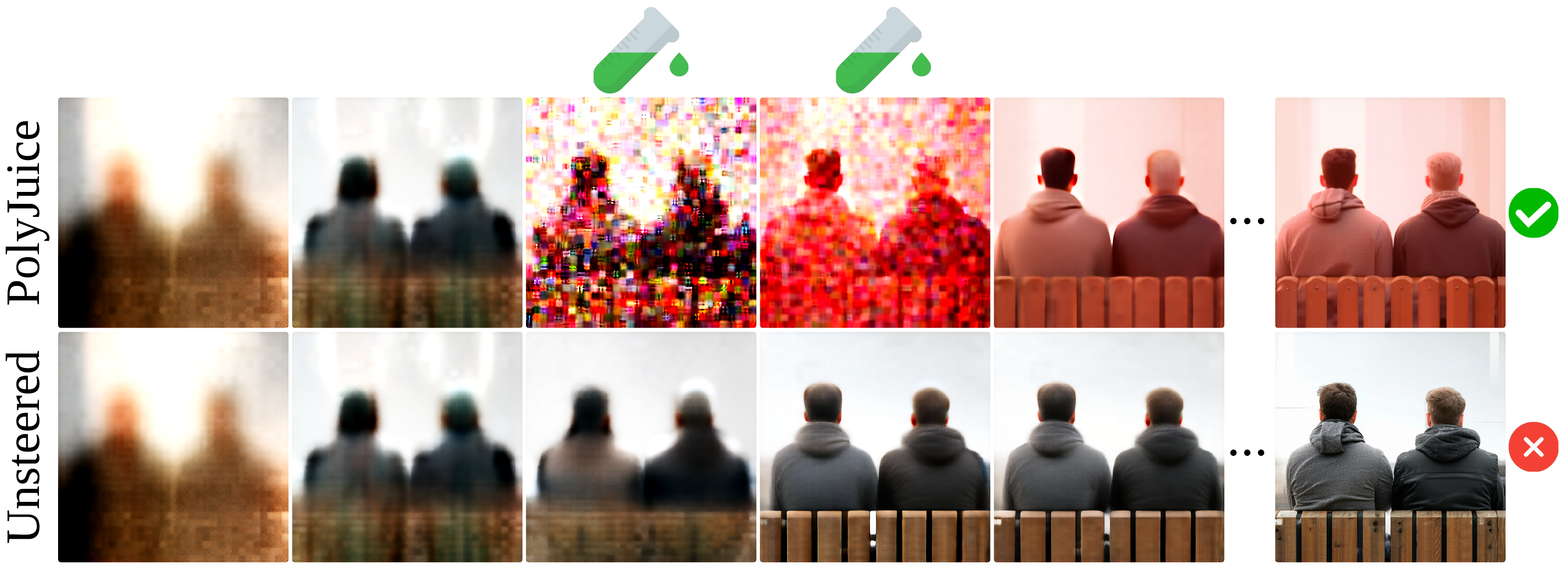}
        \caption{Intermediate steps for ``Two men sitting next to each other on a wooden bench.''}
        \label{fig:app:steer-b}
    \end{subfigure}

    \caption{Visualizing the effect of \methodName{} on the image generation process. (Top) Image that successfully deceives RINE. (bottom) Image that successfully deceives UFD.}
    \label{fig:app:steering-vis}
\end{figure}

In \cref{fig:app:steering-vis}, we show additional image-space visualizations of the effect of \methodName{}, by estimating the clean image at various timesteps. As noted in \cref{sec:app:hpsearch}, we only apply \methodName{} over a continuous interval of inference steps $[a, b] \subseteq [0, T)$, as can be seen from the figure. In both \cref{fig:app:steer-a,fig:app:steer-b}, the original unsteered T2I image generation process (bottom rows) produces an image that is detected by the SID as fake. \methodName{} steers the T2I to generate images that deceive the detectors (top rows).

\subsection{Validity of \methodName{} Across Diverse Prompts}
\label{app:reb:diverse}
 \noindent\textbf{Validity Across Diverse Prompt Categories:} To determine whether the effects of \methodName{} are applicable across a variety of categories of generated content, we inspect the COCO validation prompts associated with the attacks generated by \methodName{} and categorize the attacks according to available meta-labels. \cref{tab:r1} shows the fraction of successful attacks per prompt category. We observe that \methodName{} generally improves the success rate across all categories, demonstrating that the discovered direction is universally valid across diverse prompts and image categories. 

 \noindent\textbf{Validity Across Prompts Beyond COCO:} We also evaluate \methodName{} attacks (directions learned from COCO) on a subset of text prompts from the PartiPrompts dataset \citep{yu2022scaling}, and present the results in \cref{tab:r5}. The results demonstrate the generalizability of the \methodName{} attacks to text prompts outside COCO.
 
\begin{table}[h]
  \centering
  \caption{Success rate per prompt category in unsteered vs.\ PolyJuice (on COCO).}
  \label{tab:r1}
  \begin{tabular}{lcccc}
    \toprule
    & \multicolumn{2}{c}{UFD} & \multicolumn{2}{c}{RINE} \\
    \cmidrule(lr){2-3} \cmidrule(lr){4-5}
    & Unsteered & PolyJuice & Unsteered & PolyJuice \\
    \midrule
    Person     & 13.7   & 69.5   & 16.1   & 83.5 \\
    Animal     & 12.5   & 70.8   & 9.3    & 90.7 \\
    Food       & 14.4   & 58.1   & 11.3   & 88.8 \\
    Vehicle    & 10.0   & 73.5   & 20.1   & 79.5 \\
    Furniture  & 18.2   & 66.1   & 17.3   & 82.6 \\
    \bottomrule
  \end{tabular}
\end{table}

\begin{table}[h]
  \centering
  \caption{Attack success rate (\%) of PolyJuice on text descriptions from the PartiPrompts dataset.}
  \label{tab:r5}
  \begin{tabular}{llcc}
    \toprule
    \textbf{T2I} & \textbf{Detector} & \textbf{Unsteered} & \textbf{PolyJuice (ours)} \\
    \midrule
    \multirow{2}{*}{SD3.5}        & UFD  & 13 & 75 \textbf{(+62)} \\
                                  & RINE &  8 & 100 \textbf{(+92)} \\
    \midrule
    \multirow{2}{*}{\dev{}}     & UFD  & 78 & 96 \textbf{(+18)} \\
                                  & RINE & 42 & 86 \textbf{(+44)} \\
    \midrule
    \multirow{2}{*}{\sch{}} & UFD  & 61 & 84 \textbf{(+23)} \\
                                  & RINE & 31 & 56 \textbf{(+25)} \\
    \bottomrule
  \end{tabular}
\end{table}

\subsection{Additional SID Detectors}

We use \methodName{}-steered \sd{} against four additional SID models: NPR\citep{tan2024rethinking}, FatFormer\citep{liu2024forgery}, DRCT \citep{pmlr-v235-chen24ay}, and CoDE\citep{baraldi2024contrasting}, for further evaluating the effectiveness of PolyJuice.  The additional results are shown in \cref{tab:r3}. These results show the effectiveness of PolyJuice in improving the success rate of the attacks by 56.7\% on CoDE, 75\% on DRCT, 82\% on NPR, and 56\% on FatFormer.

\begin{table}[!h]
  \centering
  \caption{SR (\%) of unsteered SD3.5 samples vs.\ PolyJuice.}
  \label{tab:r3}
  \begin{tabular}{lcc}
    \toprule
    \textbf{Detector} & \textbf{Unsteered SD3.5} & \textbf{PolyJuice-Steered SD3.5} \\
    \midrule
    UFD            & 12.8  & 80.6 \textbf{(+68)} \\
    RINE           & 15.3  & 99.7 \textbf{(+84)} \\
    CoDE (linear)  & 43.3  & 100.0 \textbf{(+56)} \\
    DRCT (UFD)     & 25.3  & 100.0 \textbf{(+74)} \\
    NPR            & 6.0   & 87.6 \textbf{(+81)} \\
    FatFormer      & 5.5   & 62.1 \textbf{(+56)} \\
    \bottomrule
  \end{tabular}
\end{table}
\label{app:reb:sid}

\subsection{Comparison against Transferred Attacks From Diffusion-based White Box Methods \label{app:transfer}}

Transferred Attacks are an existing approach of attacking black-box models with white-box methods. First, we consider the RINE model as our target black-box model, and train a surrogate detector (a standard ResNet-50 model) to match its responses. We then extend DiffPGD \citep{xue2023diffusion} to \dev{} and \sd{}, and subsequently perform white-box attacks on the surrogate (the attacks have a 100\% success rate). The successful attacks are then `transferred' to the true black-box detector (RINE). From \cref{tab:r4}, we observe that \methodName{} has a better attack success rate than the transfer methods.

\begin{table}[h!]
\centering
\small
\caption{SR(\%) of PolyJuice vs.\ regular ($x^n$) and realistic ($x^n_0$) attack from transferred DiffPGD$^t$.}
\label{tab:r4}
\begin{tabular}{l c c}
\toprule
 & \textbf{FLUX[dev]} & \textbf{SD3.5} \\
\midrule
Unsteered & 52.4 & 15.3 \\
DiffPGD$^t$ ($x^n$) & 57.2 & 23.5 \\
DiffPGD$^t$ ($x^n_0$) & 70.1 & 32.1 \\
\textbf{PolyJuice (ours)} & \textbf{81.2} & \textbf{99.7} \\
\bottomrule
\end{tabular}
\vspace{-0.5em} %
\end{table}

\section{Potential Defense Mechanisms Against \methodName{}\label{app:def}}

One possible defense strategy against malicious usage of \methodName{} is collapsing the FN and TP clusters of the data. \methodName{} finds a subspace that is orthogonal to the direction of shift between these two clusters, while preserving the information of the target attribute (i.e., real vs. fake). Approaches from invariant representation learning~\cite{sadeghi2022on,xie2017controllable} and fairness~\cite{dehdashtian2024fairness,dehdashtian2024utilityfairness,dehdashtian2024fairerclip} can be adopted for this purpose. As an example, in the notation of U-FaTE~\cite{dehdashtian2024utilityfairness}, the target $Y$ would be real vs. fake label, while the attribute to be removed, $S$, would be TP vs. FN. 

Additionally, we advocate for detector owners to proactively use \methodName{} to generate challenging attack samples and integrate them into their training or fine-tuning data in order to make it robust to malicious usage of \methodName{}.

\end{document}